\documentclass[sigconf]{acmart}
\renewcommand\footnotetextcopyrightpermission[1]{}

\usepackage{soul}
\AtBeginDocument{%
  }

\setcopyright{acmlicensed}
\copyrightyear{2018}
\acmYear{2018}
\usepackage{booktabs}
\usepackage{multirow}
\usepackage[table,xcdraw]{xcolor}

\usepackage{graphicx}      
\usepackage{xcolor}        
\usepackage[most]{tcolorbox} 
\usepackage{hyperref}

\usepackage{algorithm}
\usepackage{subcaption}
\usepackage{algorithmic}
\newcommand{\framework}{EfficientPosterGen}
\newcommand{\Mone}{Semantic-aware Key Information Retrieval}
\newcommand{\Mtwo}{Visual-based Context Compression}
\newcommand{\Mthree}{Agentless Layout Violation Detection}
\newcommand{\Sone}{Paragraph Grouping}
\newcommand{\Stwo}{Semantic Graph Construction}
\newcommand{\Sthree}{Diversity-aware Key Segment Selection}

\usepackage{tcolorbox}
\tcbuselibrary{breakable, skins}
\usepackage{enumitem}
\usepackage{booktabs}
\usepackage{array}
\usepackage[normalem]{ulem}
\useunder{\uline}{\ul}{}

\begin{document}

% \title{\framework: Towards Token-Efficient Poster Generation from Scientific Papers}

% \title{\framework: Efficient Poster Generation from Papers via \\ Information Density-aware Retrieval and Token Compression}

\title{\framework: Semantic-aware Efficient Poster Generation via Token Compression and Accurate Violation Detection}
%%
%% Author Information
%%

\author{Wenxin Tang}
\authornote{Both authors contributed equally to this research.}
\affiliation{%
  \institution{Tsinghua University}
  \city{Beijing}
  \country{China}
}
\email{twx24@mails.tsinghua.edu.cn}

\author{Jingyu Xiao}
\authornotemark[1]
\authornote{Jingyu Xiao is the corresponding author.}
\affiliation{%
  \institution{The Chinese University of Hong Kong}
  \city{Hong Kong}
  \country{China}
}
\email{jyxiao@link.cuhk.edu.hk}

\author{Yanpei Gong}
\affiliation{%
  \institution{Harbin Institute of Technology}
  \city{Harbin}
  \country{China}
}
\email{2023211640@stu.hit.edu.cn}

\author{Fengyuan Ran}
\affiliation{%
  \institution{Wuhan University}
  \city{Wuhan}
  \country{China}
}
\email{RanFengYuanQWQ@163.com}

\author{Tongchuan Xia}
\affiliation{%
  \institution{Beijing University of Posts and Telecommunications}
  \city{Beijing}
  \country{China}
}
\email{xtc_heartune@bupt.edu.cn}

\author{Junliang Liu}
\affiliation{%
  \institution{Dalian Maritime University}
  \city{Dalian}
  \country{China}
}
\email{1120241292ljl@dlmu.edu.cn}

\author{Man Ho LAM}
\affiliation{%
  \institution{The Chinese University of Hong Kong}
  \city{Hong Kong}
  \country{China}
}
\email{mhlam@link.cuhk.edu.hk}

\author{Wenxuan Wang}
\affiliation{%
  \institution{Renmin University of China}
  \city{Beijing}
  \country{China}
}
\email{wenxuanwang@ruc.edu.cn}

\author{Michael R. Lyu}
\affiliation{%
  \institution{The Chinese University of Hong Kong}
  \city{Hong Kong}
  \country{China}
}
\email{lyu@cse.cuhk.edu.hk}

\renewcommand{\shortauthors}{Trovato et al.}

\begin{abstract}

Automated academic poster generation aims to distill lengthy research papers into concise, visually coherent presentations. Existing Multimodal Large Language Models (MLLMs) based approaches, however, suffer from three critical limitations: low information density in full-paper inputs, excessive token consumption, and unreliable layout verification. We present \framework, an end-to-end framework that addresses these challenges through semantic-aware retrieval and token-efficient multimodal generation. \framework \ introduces three core innovations: (1) Semantic-aware Key Information Retrieval (SKIR), which constructs a semantic contribution graph to model inter-segment relationships and selectively preserves important content; (2) Visual-based Context Compression (VCC), which renders selected text segments into images to shift textual information into the visual modality, significantly reducing token usage while generating poster-ready bullet points; and (3) Agentless Layout Violation Detection (ALVD), a deterministic color-gradient-based algorithm that reliably detects content overflow and spatial sparsity without auxiliary MLLMs. Extensive experiments demonstrate that \framework \ achieves substantial improvements in token efficiency and layout reliability while maintaining high poster quality, offering a scalable solution for automated academic poster generation. Our code is available at \href{https://github.com/vinsontang1/EfficientPosterGen-Code}{https://github.com/vinsontang1/EfficientPosterGen-Code}.
%   Automated academic poster generation aims to condense long research papers into concise, well-structured visual presentations. Existing MLLM-based approaches, however, suffer from low information density in full-paper inputs, excessive token consumption, and unreliable layout verification.
% We propose \framework, an end-to-end framework that addresses these limitations through information density–aware retrieval and token-efficient multimodal generation. The framework first selects high-information-density content via a structure- and semantic-aware extraction module that models contribution relationships among semantic segments. It then reduces textual token usage by embedding selected content into images and generating poster-ready bullet points from visual inputs. Finally, a deterministic color-gradient–based algorithm is introduced to verify layout validity without relying on auxiliary MLLMs.
% Experiments demonstrate that EfficientPosterGen significantly improves token efficiency and layout reliability while preserving poster quality.
\end{abstract}



\maketitle
\pagestyle{plain}

\section{Introduction}
Academic posters constitute a fundamental medium for academic communication, enabling rapid and effective dissemination of research contributions. Compared to full-length papers, posters place a stronger emphasis on information compression, structured organization, and strict layout constraints, requiring authors to distill complex ideas into concise and visually coherent representations~\cite{qiang2019learning,xu2022posterbot,chen2025posta}. Despite their importance, the manual creation of academic posters remains a labor-intensive and time-consuming process with high associated costs. However, template-based approaches~\cite{lin2023autoposter,wang2024prompt2poster} inherently impose strong rigidity in both structural and stylistic design. Such solutions struggle to flexibly adapt to the substantial variations in content distribution across different academic papers, which in turn hinders their ability to generate high-quality and expressive academic posters in realistic application scenarios.
\begin{figure}[h] 
  \centering
  \includegraphics[width=\linewidth]{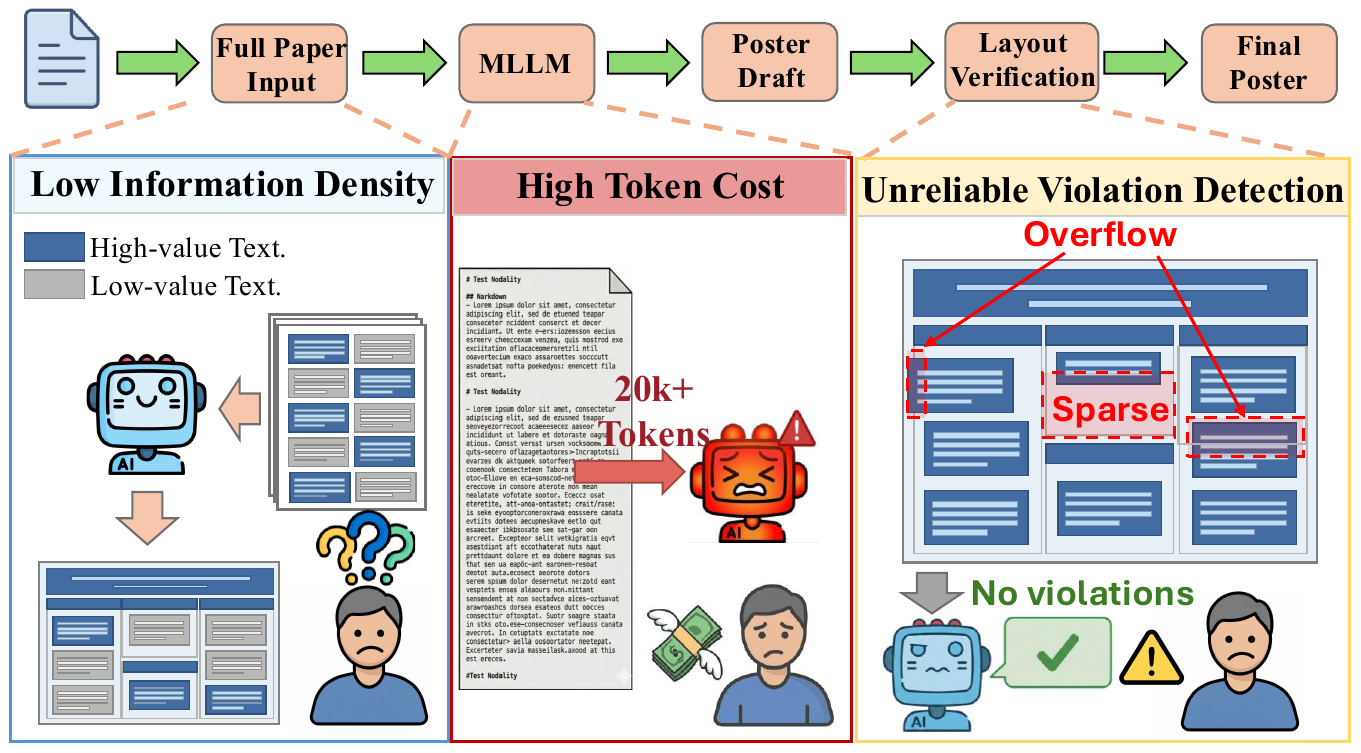}
  \caption{An overview of the workflow of existing automated poster generation approaches, along with the major challenges they encounter in practice.}
  \label{fig:intro}
\end{figure}

The rapid advancement of Multimodal Large Language Models (MLLMs) has spurred a growing body of work on multimodal content-to-code generation, including web code generation~\cite{xiao2024interaction2code, xiao2025designbench, xiao2025efficientuicoder, dang2025envisioning, wan2024mrweb, wan2025automatically}, slide code generation~\cite{tang2025slidecoder}, etc. These works demonstrate MLLMs' strong potential in producing structured, executable artifacts with code as the underlying representation. Building upon this progress, PosterAgent~\cite{pang2025paperposter} takes academic papers as input and leverages MLLMs to automatically generate editable academic posters in pptx format, enabling the automated poster generation.

% Despite PosterAgent’s capability to automatically generate academic posters from paper inputs, when facing complex and long paper documents, several significant challenges remain unaddressed.

While PosterAgent can automatically generate academic posters from paper inputs, several important challenges remain in handling complex and lengthy academic documents.

\textbf{First, the low-information-density content in academic papers tends to dilute the salient and representative information.} As illustrated on the left side of Figure~\ref{fig:intro}, a typical paper contains substantial content that is largely irrelevant to poster creation, such as references, acknowledgments, auxiliary details, and repetitive descriptions of core ideas across sections (e.g., abstract, methodology, and conclusion). In contrast, academic posters are intended to present only the most essential contributions and high-level insights. When the entire paper is provided to an MLLM in a single pass, the resulting long context makes it difficult for attention-based models~\cite{vaswani2017attention,song2025attncache,dao2022flashattention} to focus on the most critical information. Redundant and low-value content disperses attention~\cite{liu2024lost}, leading to posters that lack clear focal points or overemphasize secondary details. Moreover, the inclusion of such content significantly increases input length, further exacerbating computational overhead.

% When the entire paper is provided to an MLLM in a single pass, the inherently long context causes attention-based models to have difficulty concentrating on the most representative and semantically critical information. In particular, redundant content tends to dilute the model’s attention weights, weakening its ability to highlight core contributions and key insights. As a result, generated posters often suffer from information redundancy, lack clear focal points, and deviate from the intended emphasis of the author. From another perspective, including large amounts of low-value content also significantly increases the input token length, thereby exacerbating computational overhead with a quadratic cost under standard attention mechanisms.

\textbf{Second, directly feeding full papers to MLLMs in a pure textual modality is inherently inefficient.} A typical academic paper contains approximately 20k tokens on average. At this scale, the document length already approaches or even surpasses the maximum context window supported by many widely used large language models. For instance, Qwen3-8B and Qwen3-30B-A3B~\cite{yang2025qwen3technicalreport} support a maximum native context of 32k tokens, whereas Llama3-8B~\cite{grattafiori2024llama3herdmodels} is limited to only 8k tokens. Such excessive token inputs not only constrain model applicability due to context length limits, but also incur substantial computational and latency overhead. As a result, poster generation pipelines that rely on full-text, token-heavy inputs are difficult to scale to industrial or large-scale deployment.

\textbf{Third, layout verification with MLLMs is both costly and unreliable.} As illustrated on the right side of  Figure~\ref{fig:intro}, PosterAgent~\cite{pang2025paperposter} employs auxiliary MLLMs (e.g., painter-commenter) to provide visual feedback for detecting layout violations such as panels overflowing poster boundaries, text exceeding panel limits, or overlaps between texts across different panels. However, due to limitations in MLLMs’ element localization and spatial reasoning capabilities~\cite{liu2025benchmarking}, they often fail to accurately detect layout violations. This issue is particularly pronounced in multi-panel layouts, where MLLMs frequently struggle to identify text overflow or inter-panel overlap. Moreover, incorporating MLLMs into the layout verification pipeline introduces additional latency and token overhead.

% However, this approach delegates inherently deterministic layout constraints to probabilistic models, which introduces not only additional token overhead but also accuracy concerns. 

To address the aforementioned limitations, we introduce \framework, an end-to-end poster generation framework that reduces token costs across all stages while ensuring poster quality.

First, to filter redundant content in academic papers, we propose \textbf{\Mone\ (SKIR)}, which models a contribution graph among semantic segments. By estimating inter-segment content contribution and incorporating structural cues of the document, SKIR identifies and preserves the high-information-density semantic segments of the paper. Second, to achieve efficient token input for MLLMs while preserving semantic readability, we design the \textbf{\Mtwo \ (VCC)} ~module for poster generation, which first converts textual content into visual (image-based) representations for textual context compression and then employs an MLLM to generate concise, poster-ready bullet points from these representations. 
Third, to avoid the unreliability and token costs of MLLMs in detecting layout violations, we propose \textbf{\Mthree (ALVD)}, a color-gradient-based visual verification algorithm that deterministically checks overflow and sparse layout issues.

% This design enables more reliable verification in multi-panel cases while eliminating the additional token overhead incurred by auxiliary MLLM-based layout checking.

Overall, \framework \ enhances the poster generation process from three complementary perspectives: key content extraction, visual-based inputs representation, and output layout verification. This holistic design improves both token efficiency and layout reliability while maintaining high poster quality. Our contributions are summarized as follows:

\begin{itemize}
    \item We propose \framework, an end-to-end academic poster generation framework that reduces generation costs while maintaining high poster quality.
    \item We design \Mone, a semantic and diversity-aware content extraction strategy that models contribution relationships among semantic segments by graph to identify salient content and ensure diverse coverage of an academic paper.
    \item We introduce \Mtwo, which replaces text-based inputs with image-based representations to effectively compress long-context inputs.
    \item We develop \Mthree, a deterministic color-gradient-based layout verification algorithm that efficiently and reliably detects layout boundary violations in multi-panel posters .
\end{itemize}

% \begin{itemize}
%     \item We propose \framework, an end-to-end academic poster generation framework that reduce poster generation costs while ensuring the quality.
%     \item We design a semantic and diversity-aware content extraction strategy, \Mone, which models contribution relationships among semantic segments to identify important content and ensure diverse coverage of key ideas from an academic article.
%     \item We propose the \Mtwo, replacing the text inputs with image inputs for compressing the context.
%     \item We develop a deterministic, color-gradient-based layout verification algorithm, \Mthree, which reliably and efficiently detects layout boundary violations in multi-panel posters.
% \end{itemize}

\begin{figure*}[h]
  \centering
  \includegraphics[width=0.98\linewidth]{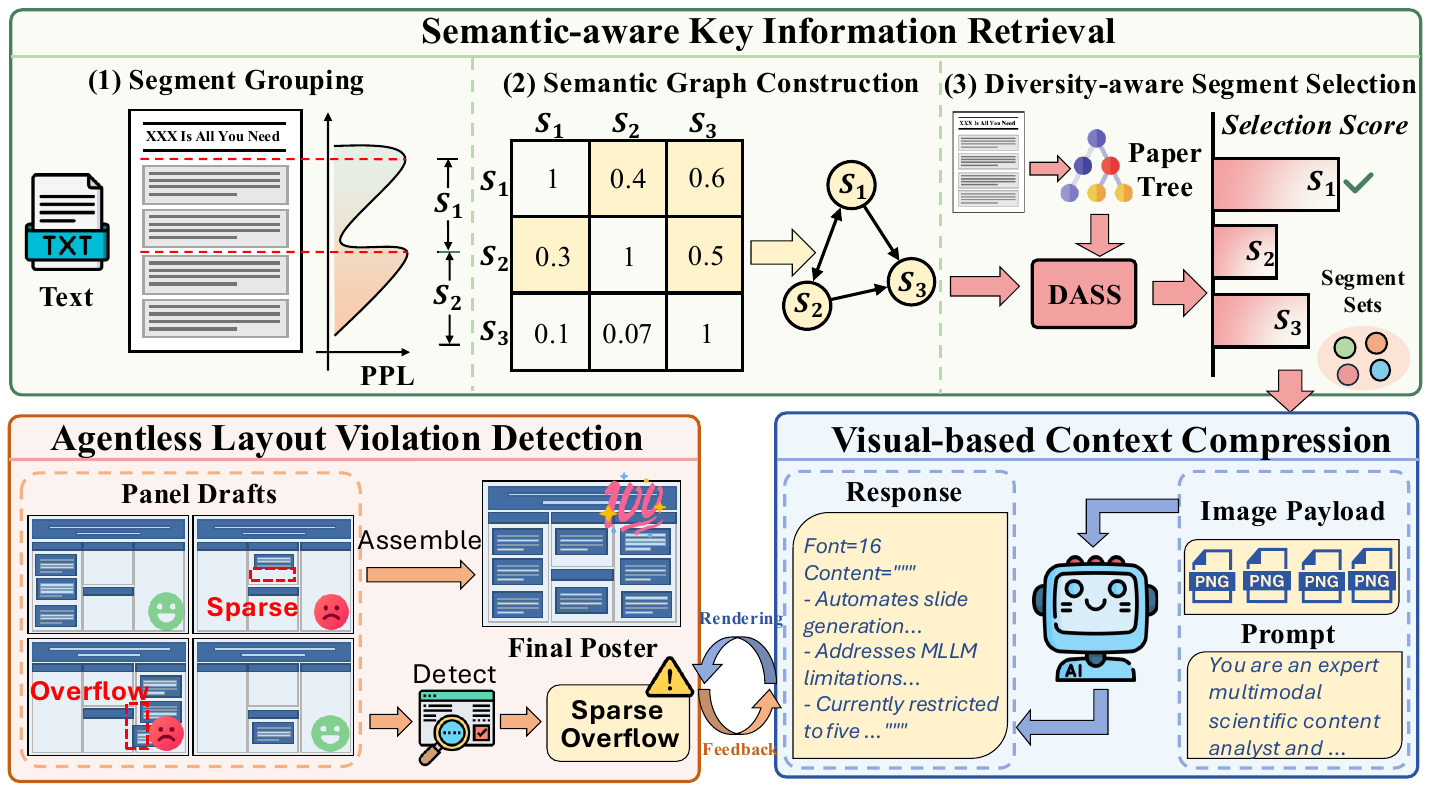}
  \caption{The framework of \framework.}\label{fig:main}
  \Description{The framework of \framework.}
\end{figure*}

\section{Background}

\subsection{Related Work}

\subsubsection{Automated Poster Generation}

Recent advances in automated academic poster generation have progressed through three key stages. Paper2Poster~\cite{pang2025paperposter} pioneers the systematic exploration of multimodal poster automation from scientific papers, establishing an end-to-end framework for this task. P2P~\cite{sun2025p2p} further advanced the field by introducing both an automated generation approach and the first fine-grained benchmark, enabling standardized evaluation and objective comparison across methods. Most recently, PosterGen~\cite{zhang2025postergen} enhances aesthetic quality through a multi-agent LLM architecture, coordinating specialized agents to improve visual appeal while maintaining information accuracy. \textit{Although these works collectively advance the field from initial feasibility to standardized evaluation and aesthetic optimization, none of them consider the efficiency of poster generation.}

\subsubsection{MLLM for Code Generation}

Multimodal Large Language Models (MLLMs) have shown strong capabilities in visually rich code generation tasks, including UI code generation~\cite{xiao2025interaction2code, xiao2025designbench, xiao2025efficientuicoder, xiao2026comuicoder, wan2025automatically, dang2025envisioning, gao2025treat}, slide generation~\cite{tang2025slidecoder}, SVG code generation~\cite{svg1,svg2,svg3,svg4}, and visually rich programming questions~\cite{rich1,rich2,rich3}. Among these, slide generation is most relevant to our work. SlideCoder~\cite{tang2025slidecoder} generates slides from reference images using a layout-aware, retrieval-augmented framework that preserves structural fidelity and produces executable slide code. PPTAgent~\cite{zheng2025pptagent} employs a two-stage, edit-based workflow guided by reference slides to ensure content quality, visual design, and structural coherence, though it relies on reference presentations rather than direct document-to-poster generation. \textit{However, existing methods rely on reference image or slides and do not address the challenges posed by long-context documents.}

% \subsubsection{Document Information Retrieval}

% However, MLLMs are not yet capable of plug-and-play use across tasks and still produce subtle errors, therefore, some studies explore their code repair abilities~\cite{repair1,repair2,repair3}.

\subsection{Task Definition}

% Given an input academic paper $P$, the objective is to automatically generate a presentation-ready academic poster in PPTX format through a multi-stage pipeline. Specifically, the process begins with an information extraction stage that distills the original paper $P$ into a refined and compact representation $P'$, where $P' = \mathcal{E}(P)$ and $\mathcal{E}(\cdot)$ denotes the information extraction operator. The refined representation preserves semantic segments with high information density while removing redundant content. Based on $P'$, a multimodal large language model (MLLM) $F$ is employed to generate a structured set of poster-ready bullet points $B$, formulated as $B = F(P')$, which serves as the high-level semantic specification of the poster content. Following the Paper2Poster setting, these bullet points are further translated into executable \texttt{python-pptx} code $C$ via a code generation function, i.e., $C = \mathcal{G}(B)$. Executing the generated code $C$ produces the final academic poster in PPTX format.

Given an input academic paper $P$, the goal is to automatically generate a presentation-ready academic poster in \texttt{PPTX} format. This task requires distilling long-form scholarly content into a compact, information-dense representation while organizing it into a structured and visually coherent poster layout. The key challenges lie in handling the long-context nature of academic papers and ensuring layout validity, as generated posters are prone to issues such as content overflow and spatial sparsity, which must be reliably detected and corrected.

\section{Methodology}

\subsection{Overview}
We present \framework, an end-to-end academic poster generation framework designed to significantly reduce token consumption throughout the pipeline. As shown in Figure~\ref{fig:main}, \framework \ is composed of three core modules that operate sequentially. First, \textbf{\Mone} (\S \ref{subsec:SKIR}) introduces an efficient input information compression strategy that jointly leverages semantic relevance and document section structure to retain  content with high information density while eliminating redundant textual segments. Second, \textbf{\Mtwo} (\S \ref{subsec:VCC}) further reduces token usage by embedding selected content segments directly into images, thereby shifting part of the textual information into the visual modality and alleviating the burden on textual input to MLLMs. Finally, \textbf{\Mthree} (\S \ref{subsec:ALVD}) provides a deterministic layout violation detection mechanism that does not rely on additional MLLMs, enabling reliable verification while reducing  token cost.

\subsection{\Mone}
\label{subsec:SKIR}

\subsubsection{\Sone}

Given an input academic paper $P$, we first employ MinerU~\cite{wang2024mineru} to parse it into three components: the textual content $T$, the media elements $M$ (e.g., figures and tables), and the hierarchical section tree $T_s$ that captures the document structure. Formally, the parsed representation is expressed as $P = \{T, M, T_s\}$. Let $t_1, t_2, \ldots, t_n \in T$ denote the individual paragraphs, where each $t_i$ is treated as the smallest atomic unit. To capture semantically coherent regions across paragraphs, it is essential to identify the boundaries where semantic transitions occur.

We employ a perplexity based method to identify semantic boundaries within the textual content. Perplexity measures how well a language model predicts a given sequence~\cite{cooper2024perplexed}; within a semantically coherent region, perplexity tends to decrease as context accumulates, whereas a sharp increase typically signals a topic shift. Suppose the current content segment begins at paragraph $t_k$ where $k \leq i$. For each paragraph $t_i$ consisting of tokens $\{w_1, w_2, \ldots, w_m\}$, we compute its perplexity conditioned on all preceding paragraphs within the current segment:

\begin{equation} \text{PPL}(t_i \mid t_{k:i-1}) = \exp\left(-\frac{1}{m}\sum_{j=1}^{m}\log P(w_j \mid t_{k:i-1}, w_{<j})\right),\end{equation}
where $t_{k:i-1} = \{t_k, t_{k+1}, \ldots, t_{i-1}\}$ denotes the paragraph sequence from index $k$ to $i-1$, $w_{<j} = \{w_1, \ldots, w_{j-1}\}$ represents all preceding tokens within the current paragraph, and $P(w_j \mid t_{k:i-1}, w_{<j})$ is the token probability estimated by a pre-trained language model. To detect semantic boundaries, we identify paragraphs where the perplexity exhibits a significant local increase relative to the preceding context. Specifically, a paragraph $t_i$ is marked as a segment boundary if the following condition holds:
\begin{equation} \text{PPL}(t_i \mid t_{k:i-1}) - \text{PPL}(t_{i-1} \mid t_{k:i-2}) > \alpha \cdot \sigma \end{equation}
where $\sigma$ denotes the standard deviation of perplexity differences computed over all consecutive paragraph pairs, and $\alpha$ is a hyperparameter controlling the sensitivity of boundary detection. When this condition is satisfied, $t_i$ is identified as the starting paragraph of a new content segment, and the segment index is updated as $k \leftarrow i$. Paragraphs satisfying this criterion typically indicate the onset of a new semantic segment, reflecting underlying thematic or structural transitions. Through this process, the paragraph sequence $T$ is partitioned into a set of content segments $\mathcal{S} = \{s_1, s_2, \ldots, s_l\}$, where each segment $s_j = \{t_{k_j}, t_{k_j+1}, \ldots, t_{k_{j+1}-1}\}$ comprises consecutive paragraphs that exhibit semantic coherence.

\subsubsection{\Stwo}

Upon obtaining the set of content segments $\mathcal{S} = \{s_1, s_2, \ldots, s_l\}$, we model the semantic relationships among segments by constructing a semantic contribution graph. Formally, we define a directed graph $G_s = (\mathcal{N}, \mathcal{E})$, where $\mathcal{N}$ denotes the node set with $|\mathcal{N}| = |\mathcal{S}|$, and each node corresponds to a content segment $s_i \in \mathcal{S}$. The edge set $\mathcal{E}$ encodes directed contribution relationships between segments.

To quantify the contribution of one segment to another, we draw inspiration from mutual information, which measures the amount of information that one variable contains about another. In our context, we aim to capture how much information segment $s_i$ provides for segment $s_j$. To this end, we compute a contribution matrix $\mathbf{X} \in \mathbb{R}^{l \times l}$ based on perplexity reduction. Intuitively, if the presence of segment $s_i$ significantly reduces the perplexity of segment $s_j$, then $s_i$ provides substantial contextual information that facilitates the prediction of $s_j$, indicating a strong semantic contribution from $s_i$ to $s_j$. Specifically, the entry $\mathbf{X}[i,j]$ is defined as:

\begin{equation}
\mathbf{X}[i,j] = \max\left(0, \frac{\text{PPL}(s_j) - \text{PPL}(s_j \mid s_i)}{\text{PPL}(s_j)}\right), \quad i \neq j,
\end{equation}
where $\text{PPL}(s_j)$ denotes the unconditional perplexity of segment $s_j$, and $\text{PPL}(s_j \mid s_i)$ denotes the perplexity of segment $s_j$ conditioned on segment $s_i$. This formulation captures the relative reduction in prediction uncertainty of $s_j$ when $s_i$ is provided as context. The $\max$ operation ensures that only positive contributions are considered, as negative values would indicate that the presence of $s_i$ increases the prediction difficulty of $s_j$, which does not constitute a meaningful semantic contribution. Additionally, we set the diagonal entries $\mathbf{X}[i,i] = 0$ for all $i$, as self-contribution is not meaningful in this context. A higher value of $\mathbf{X}[i,j]$ indicates that $s_i$ contributes more significantly to the semantic understanding of $s_j$. Based on the contribution matrix $\mathbf{X}$, we construct the directed edge set $\mathcal{E}$ according to the following condition:

\begin{equation}
(i, j) \in \mathcal{E} \iff \mathbf{X}[i,j] > \beta,
\end{equation}
where $\beta$ is a hyperparameter that controls the sparsity of the graph. That is, a directed edge from node $i$ to node $j$ is established if and only if the contribution score $\mathbf{X}[i,j]$ exceeds the threshold $\beta$.

\definecolor{vs}{RGB}{209, 242, 245}
\definecolor{hs}{RGB}{251, 244, 207}
\definecolor{ac}{RGB}{159, 250, 156}
\definecolor{bc}{RGB}{255, 183, 166}
\newcommand{\textvs}[1]{\textbf{\colorbox{vs}{#1}}}
\newcommand{\texths}[1]{\textbf{\colorbox{hs}{#1}}}
\newcommand{\textac}[1]{\textbf{\colorbox{ac}{#1}}}
\newcommand{\textbc}[1]{\textbf{\colorbox{bc}{#1}}}
\begin{figure*}[ht]
\centering
\begin{subfigure}{0.23\textwidth}
\centering
\includegraphics[width=\linewidth]{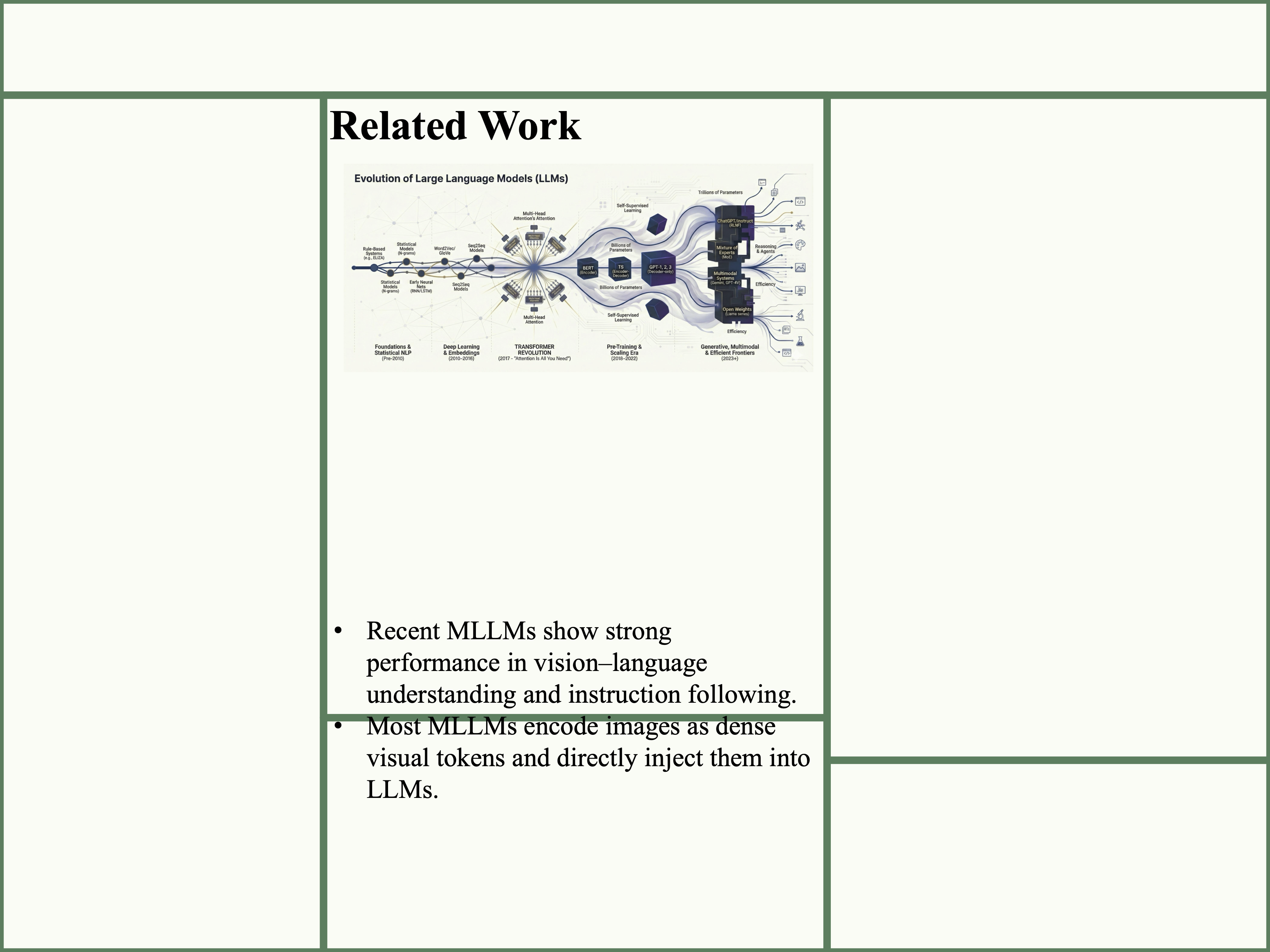}
\caption{Input Panel}
\label{fig:overflow-a}
\end{subfigure}
\hfill
\begin{subfigure}{0.23\textwidth}
\centering
\includegraphics[width=\linewidth]{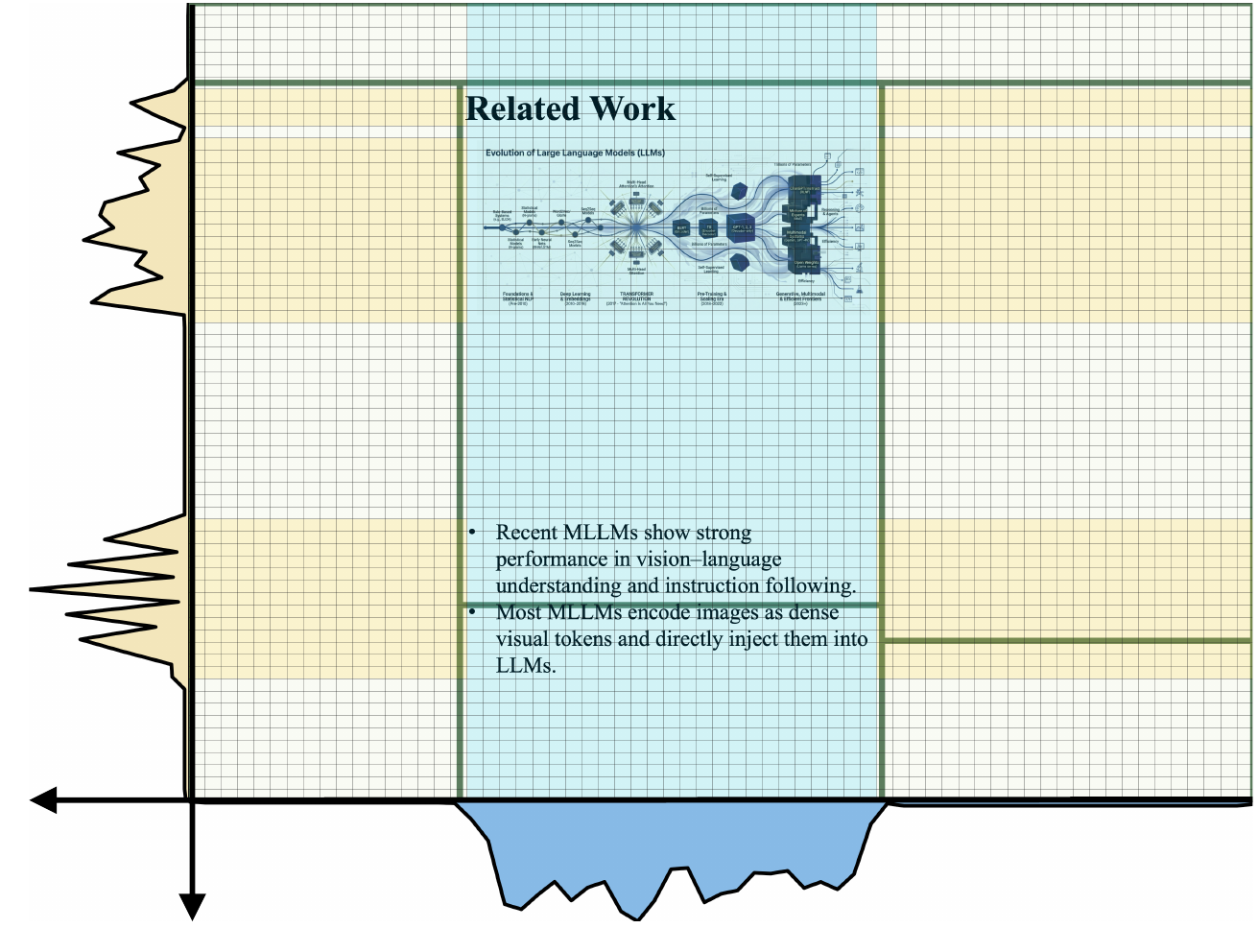}
\caption{Gradient \& Activation}
\label{fig:overflow-b}
\end{subfigure}
\hfill
\begin{subfigure}{0.23\textwidth}
\centering
\includegraphics[width=\linewidth]{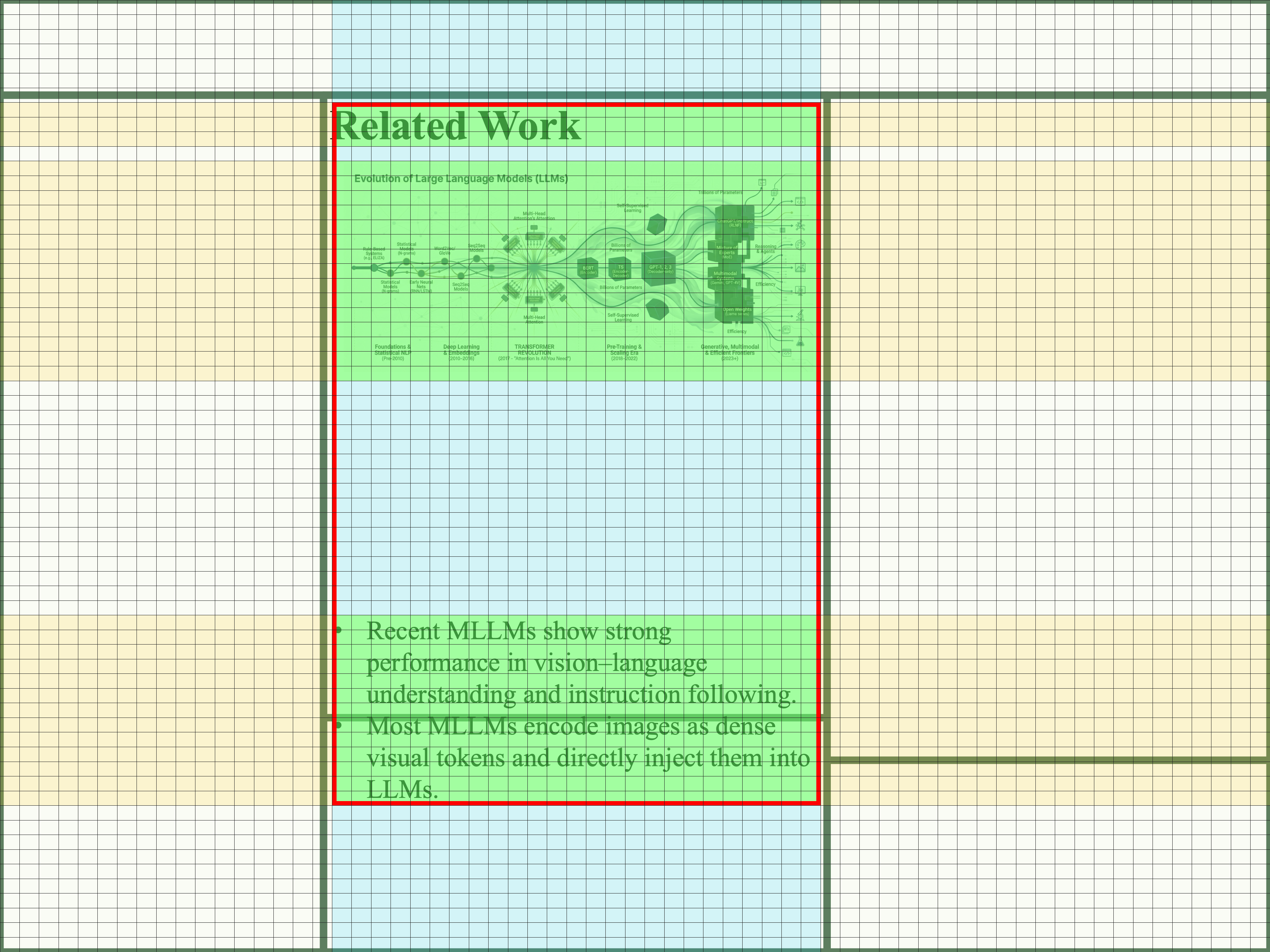}
\caption{Cartesian Product}
\label{fig:overflow-c}
\end{subfigure}
\hfill
\begin{subfigure}{0.23\textwidth}
\centering
\includegraphics[width=\linewidth]{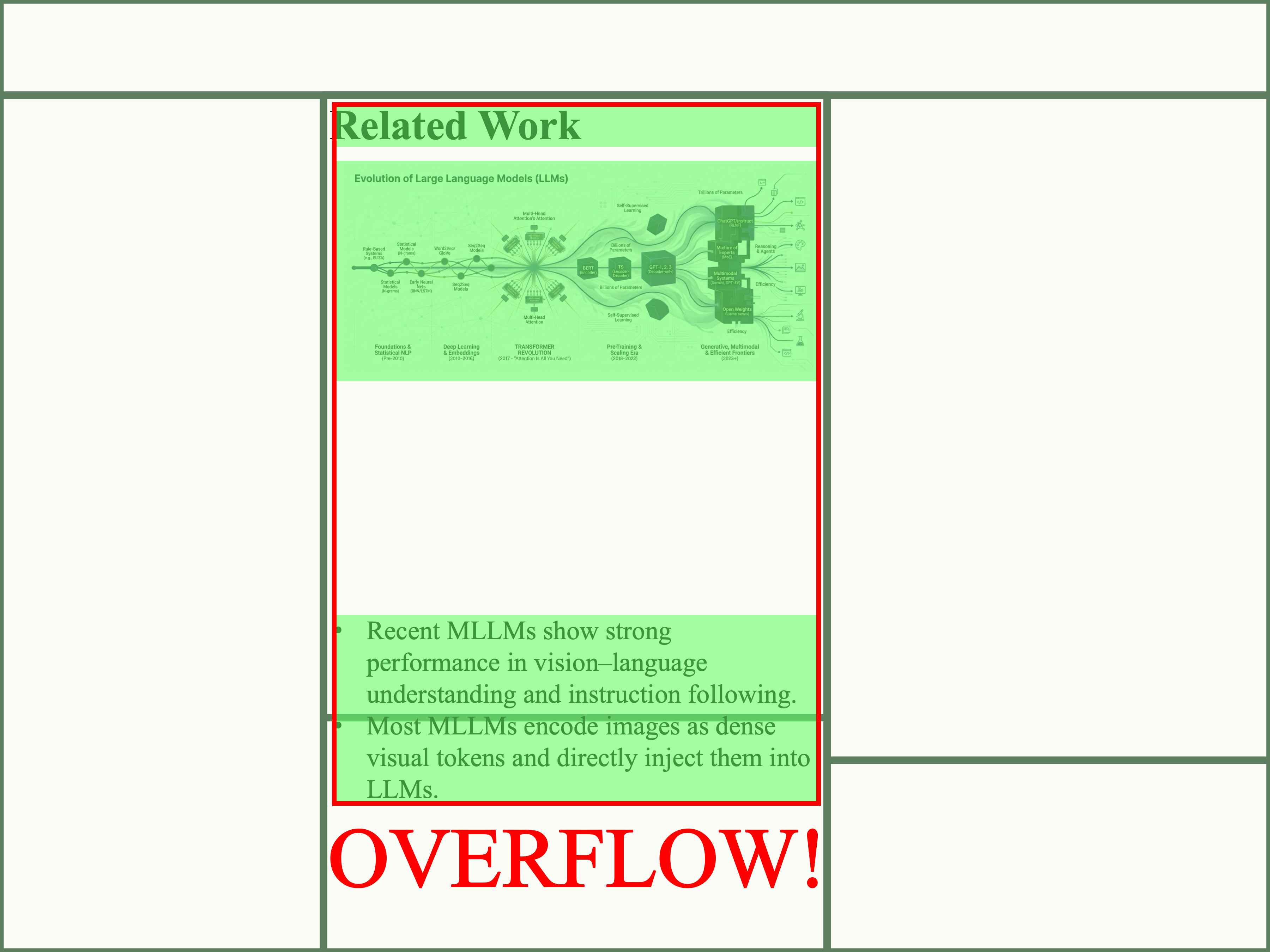}
\caption{Layout Verification}
\label{fig:overflow-d}
\end{subfigure}
\caption{An example of the layout verification process. (a) Input panel image. (b) Gradient magnitudes of vertical strips (bottom curve) and horizontal strips (left curve), with activated strips highlighted in \textvs{blue} (vertical) and \texths{yellow} (horizontal). (c) Cartesian product of activated strips yields content regions (\textac{green}) and their minimum enclosing rectangle (\textbc{red}). (d) Overflow detection via the red bounding box exceeding panel boundaries, and sparsity detection via the green coverage ratio.}
\Description{A four-panel figure illustrating the \Mthree algorithm. Panel (a) shows an input poster panel containing text and a figure. Panel (b) displays gradient magnitude curves along the bottom edge for vertical strips and along the left edge for horizontal strips, with activated vertical strips in blue and activated horizontal strips in yellow. Panel (c) shows the Cartesian product of activated strips as green regions and their minimum enclosing rectangle in red. Panel (d) demonstrates the overflow detection result, with the red bounding box and green content regions used for layout verification.}
\label{fig:overflow-example}
\end{figure*}

\subsubsection{\Sthree}

Upon constructing the semantic contribution graph $G_s = (\mathcal{N}, \mathcal{E})$, we perform content segment selection by prioritizing segments with high semantic importance (as measured by their contribution on other segments in the graph) while simultaneously ensuring semantic diversity. Notably, semantic contributions in the graph are transitive: if segment $s_i$ contributes to segment $s_j$, and $s_j$ in turn contributes to segment $s_k$, then $s_i$ implicitly plays a critical role in the information that $s_j$ provides to $s_k$, even if there is no direct edge from $s_i$ to $s_k$.

To capture this transitive influence, we employ the PageRank~\cite{gleich2015pagerank} algorithm to compute a semantic importance score for each node. PageRank is originally designed to measure the importance of web pages based on the link structure of the web. The core intuition is that a node is important if it is linked by other important nodes. In our context, since a directed edge $(i, j) \in \mathcal{E}$ indicates that segment $s_i$ contributes to segment $s_j$, we aim to assign higher importance scores to segments that contribute to many other important segments. To achieve this, we compute PageRank on the reversed graph $G_s^R = (\mathcal{N}, \mathcal{E}^R)$, where $(j, i) \in \mathcal{E}^R$ if and only if $(i, j) \in \mathcal{E}$. Formally, the PageRank score $R(s_i)$ for each segment $s_i$ is computed iteratively on the reversed graph as follows:

\begin{equation}\label{eq:pagerank}
R(s_i) = \frac{1 - d}{|\mathcal{N}|} + d \sum_{s_j \in \mathcal{N}^{+}(s_i)} \frac{R(s_j)}{|\mathcal{N}^{-}(s_j)|},
\end{equation}
where $d \in (0,1)$ is the damping factor, $\mathcal{N}^{+}(s_i)$ denotes the set of nodes that $s_i$ points to in the original graph (i.e., segments that $s_i$ contributes to), and $\mathcal{N}^{-}(s_j)$ denotes the set of nodes with edges pointing to $s_j$ in the original graph (i.e., segments that contribute to $s_j$). This formulation ensures that segments contributing to many important segments receive higher scores.

However, naively selecting segments with the highest PageRank scores via a greedy strategy may lead to a biased selection, where the chosen segments are concentrated within a limited portion of the paper, thereby lacking diverse coverage of the paper's content. 

To address this issue and ensure diversity in segment selection, we incorporate structural information from the section tree $T_s$ by introducing a penalty factor based on the Lowest Common Ancestor (LCA)~\cite{aho1973finding}. Specifically, for two segments $s_i$ and $s_j$, we define $\text{LCA}(s_i, s_j)$ as the depth of their lowest common ancestor in the hierarchical section tree $T_s$. A larger LCA depth indicates that the two segments belong to more closely related sections. Based on these considerations, we define a diversity-aware selection score that balances semantic importance and structural diversity. For a candidate segment $s_i$ and a set of already selected segments $\mathcal{S}_{\text{sel}}$, the final score $r_i$ is computed as:

\begin{equation}
r_i = R(s_i) \cdot \frac{1}{|\mathcal{S}_{\text{sel}}|} \sum_{s_j \in \mathcal{S}_{\text{sel}}} \lambda^{\text{LCA}(s_i, s_j)},
\end{equation}
where $\lambda \in (0,1)$ is a decay factor that modulates the penalty strength based on the LCA depth.

\begin{algorithm}[t]
\caption{Diversity-aware Segment Selection (DASS)}
\label{alg:selection}
\begin{algorithmic}[1]
\REQUIRE Semantic contribution graph $G_s = (\mathcal{N}, \mathcal{E})$, section tree $T_s$, compression ratio $\gamma \in (0,1)$, decay factor $\lambda$
\ENSURE Selected segment set $\mathcal{S}_{\text{sel}}$
\STATE Construct reversed graph $G_s^R = (\mathcal{N}, \mathcal{E}^R)$
\STATE Compute PageRank scores $R(s_i)$ for all $s_i \in \mathcal{N}$ on $G_s^R$ using Eq.(\ref{eq:pagerank})
\STATE Initialize selected set $\mathcal{S}_{\text{sel}} \leftarrow \emptyset$ 
\STATE Compute selection budget $K \leftarrow \lceil \gamma \cdot |\mathcal{N}| \rceil$ 
\WHILE{$|\mathcal{S}_{\text{sel}}| < K$}
    \FOR{each $s_i \in \mathcal{S} \setminus \mathcal{S}_{\text{sel}}$}
        \IF{$\mathcal{S}_{\text{sel}} = \emptyset$}
            \STATE $r_i \leftarrow R(s_i)$
        \ELSE
            \STATE $r_i \leftarrow R(s_i) \cdot \frac{1}{|\mathcal{S}_{\text{sel}}|} \sum_{s_j \in \mathcal{S}_{\text{sel}}} \lambda^{\text{LCA}(s_i, s_j)}$
        \ENDIF
    \ENDFOR
    \STATE $s^* \leftarrow \arg\max_{s_i \in \mathcal{S} \setminus \mathcal{S}_{\text{sel}}} r_i$
    \STATE $\mathcal{S}_{\text{sel}} \leftarrow \mathcal{S}_{\text{sel}} \cup \{s^*\}$
\ENDWHILE
\RETURN $\mathcal{S}_{\text{sel}}$
\end{algorithmic}
\end{algorithm}

\begin{algorithm}[t]
\small
\caption{\Mthree\ Algorithm}
\label{alg:overflow}
\begin{algorithmic}[1]
\REQUIRE Panel image $I$, panel bounding box $B_p$, number of strips $n$, sparsity threshold $\tau_s$
\ENSURE Layout status $\in \{\texttt{valid}, \texttt{overflow}, \texttt{sparse}\}$
\STATE $H, V \leftarrow$ \textsc{SplitStrips}$(I, n)$ \hfill \texttt{// horizontal and vertical strips}
\FOR{$i = 1$ \TO $n$}
    \STATE $g^h_i \leftarrow$ \textsc{GradMag}$(h_i, x)$; $g^v_i \leftarrow$ \textsc{GradMag}$(v_i, y)$
\ENDFOR
\STATE $G^h \leftarrow \{g^h_i\}_{i=1}^{n}$; $G^v \leftarrow \{g^v_i\}_{i=1}^{n}$
\STATE $\mathcal{I}_x \leftarrow \{i \mid g^v_i > \textsc{Median}(G^v)\}$ \hfill \texttt{// activated vertical strips}
\STATE $\mathcal{I}_y \leftarrow \{i \mid g^h_i > \textsc{Median}(G^h)\}$ \hfill \texttt{// activated horizontal strips}
\STATE $\mathcal{R} \leftarrow \mathcal{I}_x \times \mathcal{I}_y$ \hfill \texttt{// Cartesian product regions}
\STATE $B_c \leftarrow$ \textsc{BoundingBox}$(\mathcal{R})$ \hfill \texttt{// minimum enclosing rectangle}
\STATE $A_c \leftarrow \sum_{r \in \mathcal{R}} \textsc{Area}(r)$ \hfill \texttt{// total activated area}
\IF{$B_c \not\subseteq B_p$}
    \RETURN \texttt{overflow}
\ELSIF{$A_c / \textsc{Area}(B_p) < \tau_s$}
    \RETURN \texttt{sparse}
\ELSE
    \RETURN \texttt{valid}
\ENDIF
\end{algorithmic}
\end{algorithm}

As illustrated in Algorithm~\ref{alg:selection}, the selection process begins by constructing the reversed graph and computing the PageRank scores for all nodes (Lines 1-2). Given a compression ratio $\gamma$, the selection budget is determined as $K = \lceil \gamma \cdot |\mathcal{N}| \rceil$, representing the target number of segments to retain (Line 4). The algorithm then iteratively selects segments until the budget is reached (Lines 5-16). In each iteration, for every candidate segment not yet selected, we compute its final score $r_i$ (Lines 7-12). For the first segment, the score is simply its PageRank value (Line 8). For subsequent selections, the score is computed by multiplying the PageRank value with the mean structural diversity factor over all previously selected segments (Line 10). The segment with the highest score is then added to the selected set (Lines 14-15).

\subsection{\Mtwo \ for Poster Generation}
\label{subsec:VCC}

\subsubsection{Visual-based Context Compression} Upon obtaining the key content segments with high information density $\mathcal{S}_{\text{sel}}$, the pipeline proceeds to the MLLM-based summarization stage. We adopt a visual-based text encoding strategy to further reduce token inputs, where textual content is rendered onto images and provided to the MLLM as visual input. Prior work~\cite{wei2025deepseek,cheng2025glyph,li2025text} has demonstrated that this approach can reduce token usage by approximately 50\% while maintaining comparable model performance. Specifically, segments are grouped by their corresponding top-level sections (e.g., Introduction, Methodology) and rendered onto separate PNG images accordingly. The token compression ratio $\rho$ achieved by this visual encoding strategy is defined as:

$$
\rho = \frac{N_{\text{text}}(\mathcal{S}_{\text{sel}}) - N(\mathcal{I})}{N_{\text{text}}(\mathcal{S}_{\text{sel}})},
$$
where $N_{\text{text}}(\mathcal{S}_{\text{sel}})$ denotes the number of tokens required when the selected segments are directly provided as textual input, I is the rendered image sets and  $N(\mathcal{I})))$ denotes the token count when the rendered images are provided as visual input. A higher compression ratio indicates greater token efficiency, allowing the same content to be processed with significantly fewer tokens.

\begin{table*}[h]
\centering
\caption{Overall comparison of poster generation methods across efficiency, visual quality, and downstream evaluation metrics. \textbf{Bold} indicates the best result and \underline{underline} indicates the second-best result for each metric. API pricing details used for cost estimation are provided in Table~\ref{tab:api-pricing}.}
\label{tab:main}
\resizebox{0.8\textwidth}{!}{%
\begin{tabular}{lccccccccc}
\toprule
\multicolumn{1}{c}{} &
  \multicolumn{2}{c}{\textbf{Efficiency}} &
  \multicolumn{3}{c}{\textbf{Vis. Quality \& Txt.coherence}} &
  \multicolumn{2}{c}{\textbf{VLM-as-Judge}} &
  \multicolumn{2}{c}{\textbf{PaperQuiz}} \\ \cmidrule(l){2-10} 
\multicolumn{1}{c}{\multirow{-2}{*}{\textbf{Model}}} &
  \textbf{Token(K)$\downarrow$} &
  \textbf{Cost(\$)$\downarrow$} &
  \textbf{Vis.Sim.(\%)$\uparrow$} &
  \textbf{PPL$\downarrow$} &
  \textbf{Fig.Rel.(\%)$\uparrow$} &
  \textbf{Aesthetic$\uparrow$} &
  \textbf{Information$\uparrow$} &
  \textbf{Verbatim$\uparrow$} &
  \textbf{Interpretive$\uparrow$} \\ \midrule
\multicolumn{10}{c}{\textit{\textbf{Oracle Methods}}} \\
\rowcolor[HTML]{F8F8F8} 
Paper            & -      & -      & 53.00 & \textbf{4.60}  & 22.00 & \textbf{3.58} & \textbf{4.22} & 87.72  & 86.92 \\
\rowcolor[HTML]{F8F8F8} 
GT Poster        & -      & -      & -      & 11.26 & 21.00 & 3.56 & \underline{3.98} & 116.02 & 142.09 \\ \midrule
\multicolumn{10}{c}{\textit{\textbf{End-to-end Methods}}} \\
\rowcolor[HTML]{DCF5FF} 
5-HTML           & 26.90  & 0.1097 & 66.75 & 10.24 & 22.14 & 3.49 & 3.85 & 109.94  & 130.90 \\ \midrule
\multicolumn{10}{c}{\textit{\textbf{PosterAgent Variants}}} \\
\rowcolor[HTML]{FFEEDE} 
PosterAgent-5    & 254.37 & 0.6261 & 74.57 & 9.23  & 20.19 & 2.49 & 3.29 & 110.02  & \underline{145.63} \\
\rowcolor[HTML]{FFEEDE} 
PosterAgent-Qwen & 125.25 & \underline{0.0127} & 75.57 & 8.91 & 20.10 & 2.78 & 3.68 & 109.43 & 142.81 \\ \midrule
\multicolumn{10}{c}{\textit{\textbf{EfficientPosterGen Variants}}} \\
\rowcolor[HTML]{E4FFE4} 
Ours-5 &
  \underline{21.38} &
  0.1308 &
  \textbf{77.98} &
  \underline{8.54} &
  \underline{23.57} &
  \underline{3.57} &
  3.94 &
  \textbf{119.51} &
  \textbf{152.74} \\
\rowcolor[HTML]{E4FFE4} 
Ours-Qwen & \textbf{10.33} & \textbf{0.0016} & \underline{76.99} & 9.13 & \textbf{24.02} & 3.46 & 3.69 & \underline{116.85} & 144.34 \\ \bottomrule
\end{tabular}%
}
\end{table*}
\subsubsection{Poster Panel Generation} As shown in Figure~\ref{fig:main}, the \Mtwo\ module renders selected content segments as PNG images. The MLLM receives these images along with a task-specific prompt $\mathcal{P}$ (see Appendix~\ref{app-pmp-vga}) and generates structured bullet points with associated configuration parameters (e.g., font size) for each poster panel. This process can be formalized as:

$$
\mathcal{O} = \text{MLLM}(\mathcal{I}, \mathcal{P}),
$$
where $\mathcal{O} = \{(B_i, C_i, \Theta_i)\}_{i=1}^{N_p}$ represents the structured output for $N_p$ poster panels, with each tuple comprising bullet points $B_i$, content $C_i$, and configuration parameters $\Theta_i$ for the $i$-th panel. Following PosterAgent~\cite{pang2025paperposter}, we employ a binary-tree layout strategy~\cite{qiang2019learning} to translate the MLLM-generated bullet points into panel bounding boxes, which reliably estimates content length, maintains reading order, and preserves aspect ratios.

% \subsection{\Mtwo}

% Upon obtaining the key content segments with high information density $\mathcal{S}_{\text{sel}}$, the pipeline proceeds to the MLLM-based summarization stage. To further reduce token consumption, we adopt an visual-based text encoding strategy, where textual content is rendered onto images and provided to the MLLM as visual input. Prior work has demonstrated that this approach can reduce token usage by approximately 50\% while maintaining comparable model performance.

% As illustrated in Figure~\ref{fig:main}, the \Mtwo\ module renders selected content segments as PNG images. The MLLM receives these images along with a task-specific prompt and generates structured bullet points with associated configuration parameters (e.g., font size) for each poster panel. Following Paper2Poster, we employ a binary-tree layout strategy to translate the MLLM-generated bullet points into panel bounding boxes, which reliably estimates content length, maintains reading order, and preserves aspect ratios.

% To ensure layout validity, a feedback mechanism is incorporated. If the generated content results in overflow or excessive whitespace, the MLLM adjusts the textual content or configuration parameters accordingly until layout constraints are satisfied or the maximum number of iterations is reached.

\subsection{\Mthree}
\label{subsec:ALVD}

Initial generated poster frequently exhibits layout violations, such as text overflowing panel boundaries or panels containing insufficient content that leads to excessive unused space. Prior work~\cite{pang2025paperposter} addresses these issues by utilizing MLLMs to detect layout violations and iteratively perform layout corrections.

To mitigate the instability and token overhead of MLLM-based detection, we introduce the \Mthree \ (ALVD) module, which employs color gradient analysis to robustly identify content overflow and spatial sparsity.

As shown in Algorithm~\ref{alg:overflow}, we partition the input panel image (Figure~\ref{fig:overflow-a}) into $n$ horizontal strips $H = \{h_1, \ldots, h_n\}$ along the image height and $n$ vertical strips $V = \{v_1, \ldots, v_n\}$ along the image width (Line 1). For each strip, we compute the color gradient magnitude~\cite{zhang2017gradient} along its longitudinal direction: horizontal strips along the x-axis and vertical strips along the y-axis (Lines 2-4). As a result, content regions with substantial color variations yield high gradient values, whereas panel boundaries exhibit minimal gradients.

To identify content regions with high gradient values, we activate strips whose gradient magnitude exceeds the median value (Lines 6-7). As illustrated in Figure~\ref{fig:overflow-b}, the bottom side displays the gradient magnitudes corresponding to vertical strips, the left side shows those corresponding to horizontal strips, and the activated strips are highlighted in blue (vertical) and yellow (horizontal).

The content regions $\mathcal{R}$ are obtained by computing the Cartesian product of the activated strip indices, i.e., $\mathcal{R} = \mathcal{I}_x \times \mathcal{I}_y$ (Line 8), yielding multiple rectangular regions shown in green in Figure~\ref{fig:overflow-c}. We then compute the minimum enclosing bounding box $B_c$ of these regions (Line 9), depicted as the red rectangle in Figure~\ref{fig:overflow-c}, and the total activated area $A_c$ (Line 10). By comparing $B_c$ with the panel bounding box $B_p$ and evaluating the area ratio (Lines 11-16), the algorithm determines the layout status: \texttt{overflow} if the content exceeds panel boundaries (detected via the red bounding box), \texttt{sparse} if the coverage ratio of the green regions falls below the sparsity threshold $\tau_s$, and \texttt{valid} otherwise, as depicted in Figure~\ref{fig:overflow-d}.

% The \Mthree\  module operates on individual panels produced by the \Mtwo\  module. When an \texttt{overflow} or \texttt{sparse} status is detected, feedback containing status signal is transmitted to \Mtwo, which regenerates the content with adjusted text or configuration parameters—for instance, reducing text length or decreasing font size in the case of \texttt{overflow}, or expanding content or increasing font size in the case of \texttt{sparse}. This iterative process continues until a \texttt{valid} status is achieved or the maximum number of retries is reached. Finally, all validated panels are assembled to produce the final poster.

The \Mthree\ module operates on individual panels generated by the \Mtwo\ module. When an \texttt{overflow} or \texttt{sparse} status is detected, a corresponding status signal is fed back to \Mtwo, which regenerates the content by adjusting textual or layout parameters—for example, reducing text length or font size in the case of \texttt{overflow}, or expanding content and increasing font size in the case of \texttt{sparse}.

\section{Experiments Setup}

\subsection{Models}
We implement two variants of our proposed method with different backbone models. \textbf{Ours-5} employs GPT-5-20250807 as the backbone for both the \Mtwo\ module and internal reasoning components. \textbf{Ours-Qwen} adopts Qwen3-VL-8B-Instrcut~\cite{yang2025qwen3technicalreport} as a purely open-source alternative.

\subsection{Baselines}

We compare against four categories of baselines: \textbf{(i) Oracle methods}, including the original paper PDF (\textit{Paper}) and author-designed posters (\textit{GT Poster}). \textbf{(ii) End-to-end methods}, where GPT-5 directly generates posters through html-based rendering (\textit{5-HTML})~\cite{liu2026posterverse} \textbf{(iii) PosterAgent~\cite{pang2025paperposter}}, the first poster generate specific approach; we evaluated it on both GPT-5 (\textit{PosterAgent-5}) and Qwen3-VL-8B-Instrct (\textit{PosterAgent-Qwen}) backbones.

\subsection{Metrics}

Following Paper2Poster, we evaluate the visual and content quality of generated posters across four complementary dimensions. \textbf{(1) Visual Quality.} \textit{Visual Similarity} computes the CLIP similarity~\cite{chen2023altclip} between the generated poster and the ground-truth poster(\textit{GT Poster}), assessing whether outputs are genuinely poster-like rather than article-like layouts. \textit{Figure Relevance} computes the CLIP similarity between figures in the poster and their corresponding textual descriptions, evaluating the alignment between visual elements and textual content. \textbf{(2) Textual Coherence.} Perplexity (PPL) of the entire poster text is computed under Llama-2-7B-hf~\cite{touvron2023llama}. Lower PPL indicates more fluent and coherent language. \textbf{(3) Holistic Assessment (VLM-as-Judge).} GPT-4o assigns scores on a 1--5 scale across six criteria: three under \textit{Aesthetic Score} (Element Quality, Layout Balance, Engagement) and three under \textit{Information Score} (Clarity, Content Completeness, Logical Flow). \textbf{(4) PaperQuiz.} Using the question set provided by Paper2Poster~\cite{pang2025paperposter} (generated by GPT-o3), each poster is presented to two readers: GPT-4o mini, and GPT-o3 to answer questions based solely on the poster content. The Raw Accuracy $s_r$ is computed as the proportion of correctly matched answers. To discourage excessive verbosity, a length-based penalty is incorporated to produce an \textit{Adjusted Score} $s_a$: $s_a = s_r \left(1 + \frac{1}{\max(1, L/W)}\right)$, where $L$ denotes the total text length of the generated poster, and $W$ is the median text length of human-designed posters. Detailed metric definitions are provided in Appendix~\ref{appendix:metrics}.

\subsection{Research Questions}

To comprehensively evaluate the proposed \framework, we organize our experiments around the following research questions:

\begin{itemize}
\item \textbf{RQ1:} How does \framework\ compare to existing baselines in poster quality and token efficiency? (\S\ref{sec:main_exp})
\item \textbf{RQ2:} How do the key hyperparameters of each module affect the overall framework performance? (\S\ref{subsec-param})
\item \textbf{RQ3:} What is the individual contribution of each proposed module (SKIR, VCC, and ALVD) to the system? (\S\ref{subsec:ab_exp})
\item \textbf{RQ4:} What qualitative differences exist between posters from \framework\ and baselines? (\S\ref{subsec:casestudy})
\end{itemize}

\noindent Additional research questions addressing token consumption analysis, comparison with multi-agent approaches, layout detection comparison, and human evaluation are provided in Appendix~\ref{appendix:rq}.

\section{Experimental Results}

\begin{figure}[h] 
  \centering
  \includegraphics[width=0.8\linewidth]{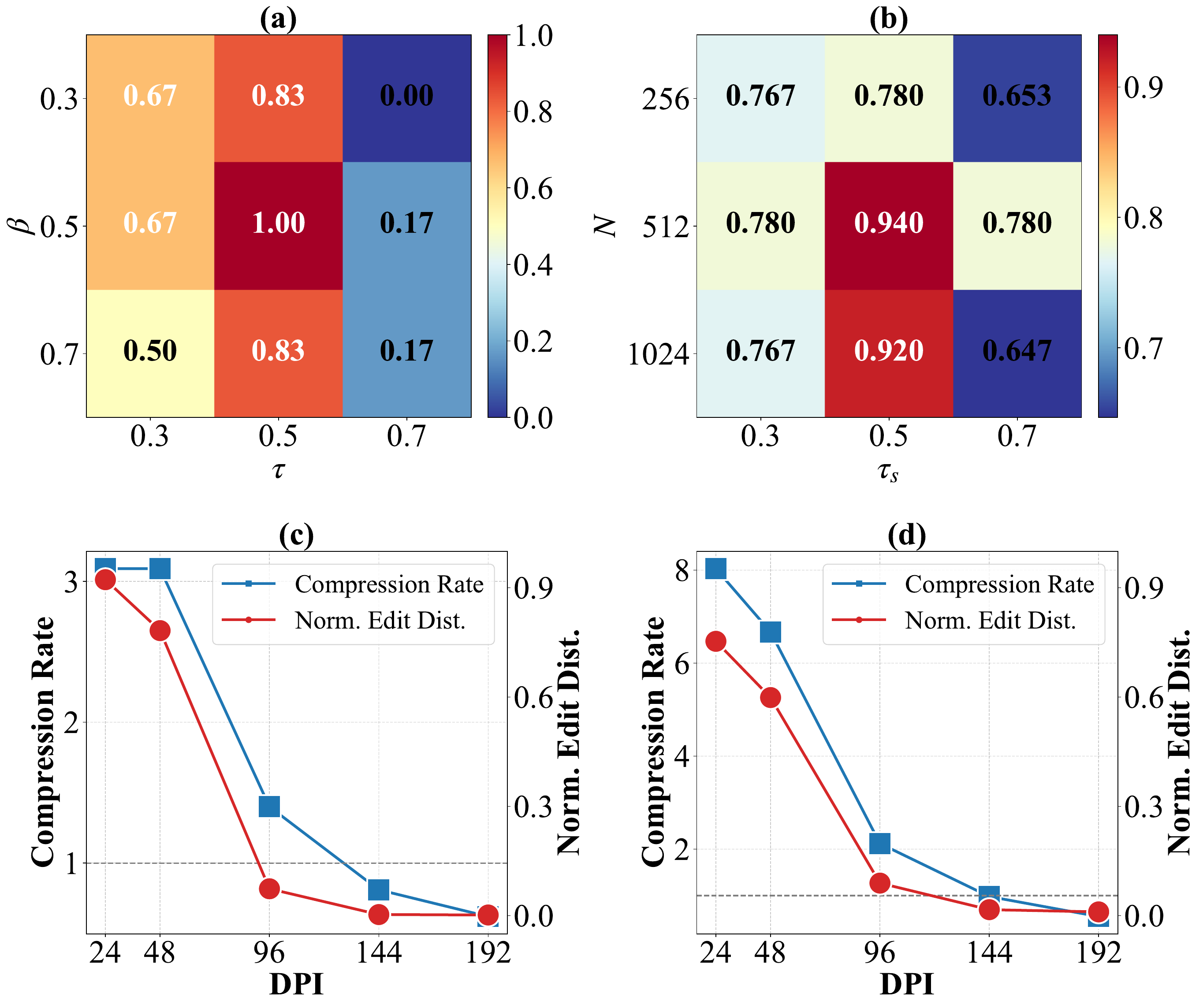}
\caption{Parameter sensitivity analysis. 
(a) Entropy reduction ratio under varying $\beta$ and $\gamma$. 
(b) Layout detection accuracy across strip numbers $N$ and activation thresholds $\tau_s$. 
(c)-(d) Compression ratio vs. normalized edit distance at varying DPI for GPT-5 and Qwen3-VL-8B-Instruct.}

  \label{fig:param-study}
\end{figure}
\begin{figure*}[ht] 
  \centering
  \includegraphics[width=0.76\linewidth]{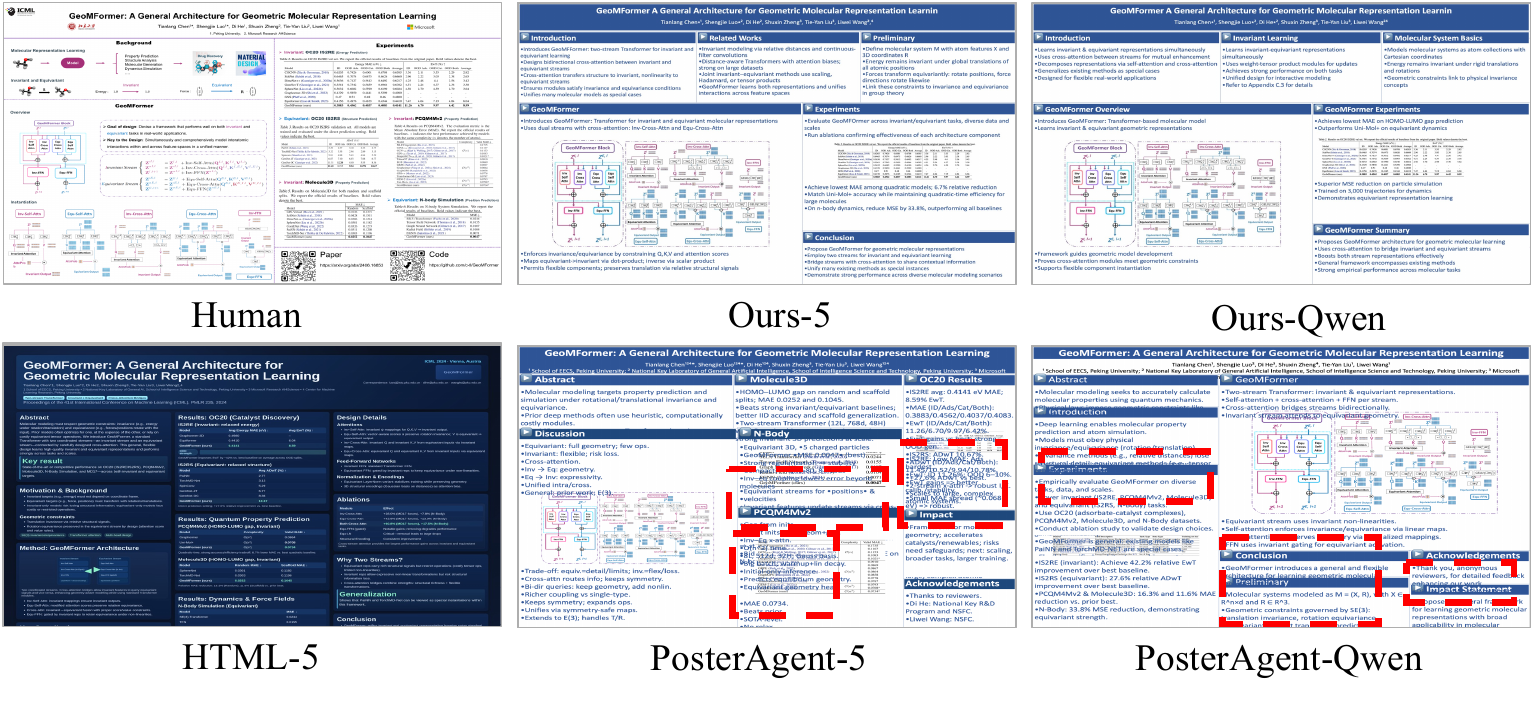}
  \caption{Examples of posters generated by different methods.}
  \label{fig:casestudy}
\end{figure*}

\subsection{Quantitative Results and Analysis}
\label{sec:main_exp}

Table~\ref{tab:main} presents the overall performance of different frameworks across all evaluation dimensions, with detailed evaluation results provided in Appendix~\ref{app-detail_main}. Our framework adopts the optimal hyperparameter configuration identified in the parameter study (Section~\ref{subsec-param}). As shown in the table, \framework\ achieves the best or second-best results on the majority of evaluation metrics while requiring substantially fewer tokens. Specifically, the Ours-5 and Ours-Qwen variants complete the entire paper-to-poster pipeline using only 21.38K and 10.33K tokens, respectively, which represents a nearly 10$\times$ reduction compared to PosterAgent. Notably, the majority of PosterAgent's token overhead stems from its layout validity verification stage (see Appendix~\ref{app-main-eff}), which relies on MLLM calls to assess layout compliance; in contrast, the deterministic, agentless design of \Mthree\ (ALVD) eliminates the need for auxiliary MLLM calls during layout verification, thereby incurring zero additional token cost. Moreover, by replacing probabilistic MLLM-based judgments with a deterministic algorithm, \Mthree\ substantially reduces false positive rates in layout violation detection, which in turn avoids unnecessary panel-level content regeneration cycles and the associated token overhead that such cycles would otherwise incur. Furthermore, owing to the efficient information extraction capability of \Mone\ (SKIR), which identifies and retains the most key semantic segments within a constrained token budget, our framework achieves strong performance on the PaperQuiz evaluation, where the GPT-5 and Qwen variants of \framework\ obtain overall scores of 119.51 and 152.74, surpassing PosterAgent by 9.49 and 7.11 points, respectively.

\subsection{Parameter Study}
\label{subsec-param}
% \begin{figure*}[h]
% \centering
% \begin{subfigure}{0.23\textwidth}
% \centering
% \includegraphics[width=\linewidth]{figures/param_m3.pdf}
% \caption{Input Panel}
% \label{fig:overflow-a}
% \end{subfigure}
% \hfill
% \begin{subfigure}{0.23\textwidth}
% \centering
% \includegraphics[width=\linewidth]{figures/param_m3.pdf}
% \caption{Gradient \& Activation}
% \label{fig:overflow-b}
% \end{subfigure}
% \hfill
% \begin{subfigure}{0.23\textwidth}
% \centering
% \includegraphics[width=\linewidth]{figures/param_m3.pdf}
% \caption{Cartesian Product}
% \label{fig:overflow-c}
% \end{subfigure}
% \hfill
% \begin{subfigure}{0.23\textwidth}
% \centering
% \includegraphics[width=\linewidth]{figures/param_m3.pdf}
% \caption{Layout Verification}
% \label{fig:overflow-d}
% \end{subfigure}
% \caption{ddd}
% \Description{ddd}
% \label{fig:overflow-example}
% \end{figure*}

To systematically evaluate each module within \framework, we design three sub-tasks that isolate the performance of \Mone, \Mtwo, and \Mthree, respectively. Detailed configurations and results are provided in the Appendix~\ref{appendix:param}.

\begin{table}[]
\centering
\caption{Overall performance of ablation study.}
\label{tab:ab}
\resizebox{0.95\linewidth}{!}{\begin{tabular}{ccccc}
\hline
                   & \multicolumn{2}{c}{\textbf{Efficiency}} & \multicolumn{1}{l}{} & \multicolumn{1}{l}{} \\ \cline{2-3}
\multirow{-2}{*}{\textbf{Setting}} &
  \textbf{Token(K)$\downarrow$} &
  \textbf{Cost(\$)$\downarrow$} &
  \multicolumn{1}{l}{\multirow{-2}{*}{\textbf{VLM-as-Judge$\uparrow$}}} &
  \multicolumn{1}{l}{\multirow{-2}{*}{\textbf{PaperQuiz$\uparrow$}}} \\ \hline
\rowcolor[HTML]{F8F8F8} 
EfficientPosterGen & \textbf{10.35}     & \textbf{0.0016}    & \textbf{3.64}        & \textbf{133.35}      \\
\rowcolor[HTML]{F8F8F8} 
w/o SKIR           & 12.10              & 0.0018             & 3.53                 & 129.33               \\
\rowcolor[HTML]{F8F8F8} 
w/o VCC            & 14.48              & 0.0020             & 3.59                 & 131.32               \\
\rowcolor[HTML]{F8F8F8} 
w/o ALVD           & 84.78              & 0.0076             & 3.27                 & 125.82               \\ \hline
\end{tabular}}
\end{table}

\textbf{Task 1:} We adopt an entropy-based metric to quantify the representativeness of selected segments. Let $H(P)$ denote the entropy of the full paper and $H(P \mid \mathcal{S}_{\text{sel}})$ denote the conditional entropy given the selected set $\mathcal{S}_{\text{sel}}$. To measure the average contribution of each selected token to entropy reduction, we define the normalized entropy reduction ratio as:
$
\Delta H_{\text{norm}} = \frac{H(P) - H(P \mid \mathcal{S}_{\text{sel}})}{H(P) \cdot |\mathcal{S}_{\text{sel}}|}
$
where $|\mathcal{S}_{\text{sel}}|$ denotes the total token count of selected segments. For visualization, we apply min-max normalization~\cite{patro2015normalization} across all configurations to obtain comparable scores in the heatmap. We evaluate this metric under varying edge activation thresholds $\beta$ and segment budgets $\gamma$, which govern the graph sparsity and selection scale, respectively.
As shown in Figure~\ref{fig:param-study}(a), the configuration $\beta=0.5$ with $\gamma=0.5$ achieves the highest normalized score, indicating optimal per-token efficiency in entropy reduction. Notably, larger segment budgets ($\gamma=0.7$) lead to diminished efficiency across all $\beta$ settings, as excessive selection introduces redundant content that contributes marginally to information coverage while inflating the token count.

\textbf{Task 2:} We formulate an OCR-based task to examine the trade-off between recognition accuracy and token consumption. The content segments are rendered at varying DPI settings, from which an OCR model recovers the original text. We evaluated it on GPT-5 and Qwen3-VL-8B-Instruct, adopting normalized edit distance~\cite{ristad2002learning} and compression ratio as the evaluation metric, which quantifies the fidelity of visual representations by measuring the discrepancy between recovered and original content.As shown in Figure~\ref{fig:param-study}(c,d), lower DPI settings achieve higher compression ratios at the cost of increased edit distance, while higher DPI settings yield near-perfect recognition but incur greater token consumption. Notably, at DPI=96, GPT-5 (Figure~\ref{fig:param-study}(c)) and Qwen3-VL-8B-Instruct (Figure~\ref{fig:param-study}(d)) achieve compression ratios of 1.4$\times$ and 2.12$\times$ respectively, while maintaining normalized edit distances of only 7.4\% and 8.9\%.

\textbf{Task 3:} We construct a ternary classification dataset (\texttt{overflow}, \texttt{sparse}, \texttt{valid}) for layout detection, whose construction details are described in the Appendix~\ref{app-subsubdataset}. We evaluate \Mthree\ on this dataset by varying the number of strips $N$ and the activation threshold $\tau_s$, which govern the granularity and sensitivity of content region detection, respectively. 
As shown in Figure~\ref{fig:param-study}(b), $N=512$ with $\tau_s=0.5$ achieves the highest accuracy of 0.94, where $N$ governs detection granularity and $\tau_s$ controls alignment with human visual perception. Too small $N$ (256) fails to capture fine-grained boundaries, while excessively large $N$ (1024) introduces noise that undermines detection stability. $\tau_s$ values that are too high (0.7) misclassify reasonable layouts as sparse, deviating from human preferences for content density. A comparative analysis against other detection methods is in Appendix~\ref{app-color-compare}.

\subsection{Ablation Study}
\label{subsec:ab_exp}
We design three ablation settings to validate the effectiveness of individual components in our framework: (1) \textit{w/o SKIR}, which directly feeds the entire paper content without key information retrieval; (2) \textit{w/o VCC}, which provides selected content as raw text tokens rather than embedded images; and (3) \textit{w/o ALVD}, which replaces the deterministic verification algorithm with an MLLM-based approach. Detailed descriptions and results are provided in Appendix~\ref{appendix:ablation-detail}. We randomly sample 20\% of the test instances and employ Qwen3-VL-8B-Instruct as the backbone model. The overall results are presented in Table~\ref{tab:ab}, where removing each component leads to varying degrees of degradation in both evaluation scores and token efficiency. Notably, the \textit{w/o ALVD} setting exhibits a substantial increase in token consumption (84.78K), which demonstrates that our proposed agentless algorithm effectively controls the token budget while avoiding unnecessary regeneration cycles.

\subsection{Case Study}
\label{subsec:casestudy}
Figure~\ref{fig:casestudy} presents a qualitative comparison across different methods on a representative case, with additional examples in the Appendix~\ref{app-casestudy}. Specifically, we showcase the human-designed ground truth poster alongside outputs generated by our \framework\ and PosterAgent, each instantiated with two backbone models: GPT-5 and Qwen3-VL-8B-Instruct. We also include results from the end-to-end HTML-based generation approach powered by GPT-5. For fair comparison, we apply the same template style to both PosterAgent and our method. The HTML-based approach tends to produce posters with excessive word counts, resulting in considerably small font sizes that compromise human readability and accessibility.

Benefiting from the deterministic layout detection algorithm of \Mthree, posters generated by \framework\ exhibit no content overflow beyond panel boundaries, whereas PosterAgent produces layouts where content exceeds the designated panel regions (highlighted by red bounding boxes) and even extends beyond the overall poster boundary in some cases.

\section{Conclusion}
We presented \framework\, an end-to-end framework for automated academic poster generation that addresses the low information density, excessive token consumption, and unreliable layout verification inherent in existing MLLM-based approaches. Through three tightly integrated modules, SKIR, VCC, and ALVD, our framework achieves substantial improvements in token efficiency and layout reliability while maintaining high poster quality. Extensive experiments consistently demonstrate the effectiveness of \framework\ across different backbone models and evaluation settings while maintaining low token consumption.

\bibliographystyle{ACM-Reference-Format}
\bibliography{sample-base}


\appendix

\section{Additional Research Questions}
\label{appendix:rq}

In addition to the research questions addressed in the main text, we investigate the following supplementary questions to provide a more thorough evaluation of \framework:

\begin{itemize}
\item \textbf{RQ5:} How is the token consumption composed across textual and visual modalities for both input and output? (Appendix~\ref{appendix:efficiency})
\item \textbf{RQ6:} Does \framework\ generalize effectively across different backbone models, including comparisons with multi-agent approaches? (Appendix~\ref{appendix:multi-agent})
\item \textbf{RQ7:} How does the deterministic ALVD algorithm compare against alternative layout detection methods? (Appendix~\ref{app-color-compare})
\item \textbf{RQ8:} Are the improvements observed in automated metrics consistent with human judgments? (Appendix~\ref{appendix:human-eval})
\end{itemize}

\section{Efficiency Analysis}
\label{appendix:efficiency}

This appendix provides a detailed breakdown of token consumption for both the main experiment and the ablation study, offering deeper insights into the efficiency characteristics of each method and module.

\subsection{Main Experiment Token Analysis}
\label{app-main-eff}
Table~\ref{tab:efficiency-detail} reports the fine-grained token consumption of each method, decomposed into textual and visual modalities for both input and output.

The most striking observation is the high visual input token consumption of PosterAgent variants, with PosterAgent-5 consuming 187.67K and PosterAgent-Qwen consuming 82.16K visual input tokens. This overhead is predominantly attributable to the MLLM-based layout verification stage, which requires rendering the poster as an image and feeding it back to the MLLM for compliance assessment at each iteration. This finding demonstrates that MLLM-based layout detection not only suffers from unreliable probabilistic judgments, as discussed in the main text, but also introduces a substantial token burden that dominates the overall consumption.

In contrast, \framework variants maintain minimal visual input token usage at 3.70K and 2.58K for Ours-5 and Ours-Qwen, respectively, which stems solely from the image-embedded content used in the VCC module.

Regarding textual tokens, \framework variants achieve significantly lower input consumption at approximately 5.8K, compared to over 31K for PosterAgent variants. On the output side, Ours-5 exhibits a notably higher textual output of 11.90K compared to 1.88K for Ours-Qwen. This discrepancy arises because GPT-5 incorporates an internal chain-of-thought reasoning process whose thinking tokens are counted as part of the output, whereas Qwen3-VL-8B does not employ such a mechanism.

\subsection{Ablation Token Analysis}

Table~\ref{tab:ablation-efficiency-detail} presents the token consumption breakdown for each ablation setting, which isolates the efficiency contribution of individual modules.

\textbf{w/o SKIR.} Removing SKIR increases the visual input tokens from 2.56K to 4.08K, representing approximately a 1.59$\times$ increase. Without targeted segment selection, a larger volume of content is passed to the VCC module for image embedding, which directly inflates the visual token count. The textual input remains comparable, as the text-based prompt structure is largely unchanged; however, the total token consumption rises from 10.41K to 12.10K due to the expanded visual input.

\textbf{w/o VCC.} Bypassing VCC eliminates visual input tokens entirely, as all content is conveyed through the textual modality. Consequently, the textual input surges from 5.94K to 12.48K, yielding a total of 14.44K. Compared to the full pipeline, the additional textual input introduced by removing VCC amounts to 6.54K (12.48K $-$ 5.94K), which would have been compressed into only 2.56K visual tokens by the VCC module, reflecting a compression ratio of approximately 1.57$\times$.

\textbf{w/o ALVD.} Replacing the deterministic ALVD with an MLLM-based approach causes the most dramatic efficiency degradation. Visual input tokens surge from 2.56K to 71.22K—a nearly 28$\times$ increase—as each layout verification iteration requires the rendered poster image to be re-encoded and submitted to the MLLM. This massive overhead, combined with false positive detections that trigger redundant panel-level regeneration cycles, drives the total token consumption to 84.90K, which is over 8$\times$ that of the full pipeline. Notably, the textual input also nearly doubles from 5.94K to 11.76K, as each regeneration cycle reintroduces the panel-level prompts and instructions.

\begin{table*}[t]
\caption{Comparison of poster generation methods on visual quality, text coherence, and VLM-as-Judge evaluation using GPT-4o as the backbone. \textbf{Bold} denotes the best result in each column.}
\label{tab:4o-comparison}
\resizebox{\textwidth}{!}{%
\begin{tabular}{lccc|cccc|cccc|c}
\toprule
\multicolumn{1}{c}{} &
  \multicolumn{3}{c|}{\textbf{Vis. Quality \& Txt.coherence}} &
  \multicolumn{4}{c|}{\textbf{Aesthetic$\uparrow$}} &
  \multicolumn{4}{c|}{\textbf{Information$\uparrow$}} &
  \\ \cmidrule(lr){2-4} \cmidrule(lr){5-8} \cmidrule(lr){9-12}
\multicolumn{1}{c}{\multirow{-2}{*}{\textbf{Model}}} &
  \textbf{Vis.Sim.(\%)$\uparrow$} &
  \textbf{PPL$\downarrow$} &
  \textbf{Fig.Rel.(\%)$\uparrow$} &
  \textbf{Element} &
  \textbf{Layout} &
  \textbf{Engage.} &
  \textbf{Avg} &
  \textbf{Clarity} &
  \textbf{Content} &
  \textbf{Logic} &
  \textbf{Avg} &
  \multirow{-2}{*}{\textbf{Overall$\uparrow$}} \\ \midrule
\rowcolor[HTML]{F8F8F8} 
OWL-4o      & 54.00          & 11.46          & -              & 2.76          & 3.62          & 2.56          & 2.98          & 3.92          & 2.89          & 3.36          & 3.39          & 3.19 \\
\rowcolor[HTML]{F8F8F8} 
PPTAgent-4o & 50.00          & \textbf{6.20}  & 16.00          & 2.49          & 3.05          & 2.45          & 2.66          & 2.05          & 1.26          & 1.38          & 1.56          & 2.11 \\
\rowcolor[HTML]{E4FFE4} 
Ours-4o     & \textbf{75.23} & 8.96           & \textbf{23.87} & \textbf{3.92} & \textbf{3.71} & \textbf{2.93} & \textbf{3.52} & \textbf{4.09} & \textbf{3.41} & \textbf{3.66} & \textbf{3.72} & \textbf{3.62} \\ \bottomrule
\end{tabular}%
}
\end{table*}

\begin{table*}[t]
\caption{Detailed PaperQuiz evaluation results using GPT-4o as the backbone. Raw Accuracy and Density-Augmented Score are reported for both verbatim and interpretive questions, further broken down by open-source and closed-source reader models. \textbf{Bold} denotes the best result in each column.}
\label{tab:4o-paperquiz-detail}
\resizebox{\linewidth}{!}{%
\begin{tabular}{lcccccccccccc}
\toprule
\multicolumn{1}{c}{} &
  \multicolumn{7}{c}{\textbf{Raw Accuracy$\uparrow$}} &
  \multicolumn{5}{c}{\textbf{Density-Augmented Score$\uparrow$}} \\ \cmidrule(lr){2-8} \cmidrule(l){9-13}
\multicolumn{1}{c}{} &
  \multicolumn{3}{c}{\textbf{Verbatim}} &
  \multicolumn{3}{c}{\textbf{Interpretive}} &
  &
  & & \\ \cmidrule(lr){2-4} \cmidrule(lr){5-7}
\multicolumn{1}{c}{\multirow{-3}{*}{\textbf{Model}}} &
  \textbf{open-source} &
  \textbf{closed-source} &
  \textbf{V-Avg} &
  \textbf{open-source} &
  \textbf{closed-source} &
  \textbf{I-Avg} &
  \multirow{-2}{*}{\textbf{Overall}} &
  \multirow{-2}{*}{\textbf{V-Avg}} &
  \multirow{-2}{*}{\textbf{I-Avg}} &
  \multirow{-2}{*}{\textbf{Overall}} \\ \midrule
\rowcolor[HTML]{F8F8F8} 
OWL-4o      & 47.87 & 31.96 & 39.92 & 49.94 & 74.38 & 62.16 & 51.04 & 78.69 & 122.91 & 100.80 \\
\rowcolor[HTML]{F8F8F8} 
PPTAgent-4o & 39.63 & 11.99 & 25.81 & 36.22 & 37.15 & 36.68 & 31.25 & 51.62 & 73.37 & 62.49 \\
\rowcolor[HTML]{E4FFE4} 
Ours-4o     & \textbf{57.40} & \textbf{50.63} & \textbf{54.02} & \textbf{53.56} & \textbf{79.50} & \textbf{66.53} & \textbf{60.27} & \textbf{107.99} & \textbf{132.96} & \textbf{120.47} \\ \bottomrule
\end{tabular}%
}
\end{table*}

\section{Comparison with Multi-Agent Approaches}
\label{appendix:multi-agent}

To further evaluate the generalizability of \framework, we compare it against two representative multi-agent poster generation methods, OWL~\cite{hu2025owl} and PPTAgent~\cite{zheng2025pptagent}, using GPT-4o as the shared backbone. The PaperQuiz evaluation employs six reader models spanning both open-source and closed-source categories: the open-source readers include LLaVA-OneVision-Qwen2-7b-ov-hf~\cite{li2024llava}, Phi-4-multimodal-instruct~\cite{abouelenin2025phi}, and Llama-4-Scout-17B-16E-Instruct, while the closed-source readers include Gemini-2.0-Flash, GPT-4o-mini, and GPT-o3. Table~\ref{tab:4o-comparison} and Table~\ref{tab:4o-paperquiz-detail} present the results on visual quality, VLM-as-Judge, and PaperQuiz evaluations, respectively.

As shown in the tables, Ours-4o achieves the best performance on the majority of metrics across all evaluation dimensions. In terms of visual quality, Ours-4o attains a visual similarity of 75.23\% and a figure relevance of 23.87\%, substantially outperforming both baselines. For VLM-as-Judge evaluation, Ours-4o leads in both aesthetic and informational quality, achieving an overall score of 3.62 compared to 3.19 for OWL-4o and 2.11 for PPTAgent-4o. The PaperQuiz results further confirm this advantage, where Ours-4o obtains the highest raw accuracy and density-augmented scores on both verbatim and interpretive questions across open-source and closed-source reader models. These results demonstrate that \framework\ generalizes effectively to different backbone models and consistently outperforms multi-agent approaches that rely on complex inter-agent coordination.

\begin{figure}[t]
\centering
\includegraphics[width=1.0\columnwidth]{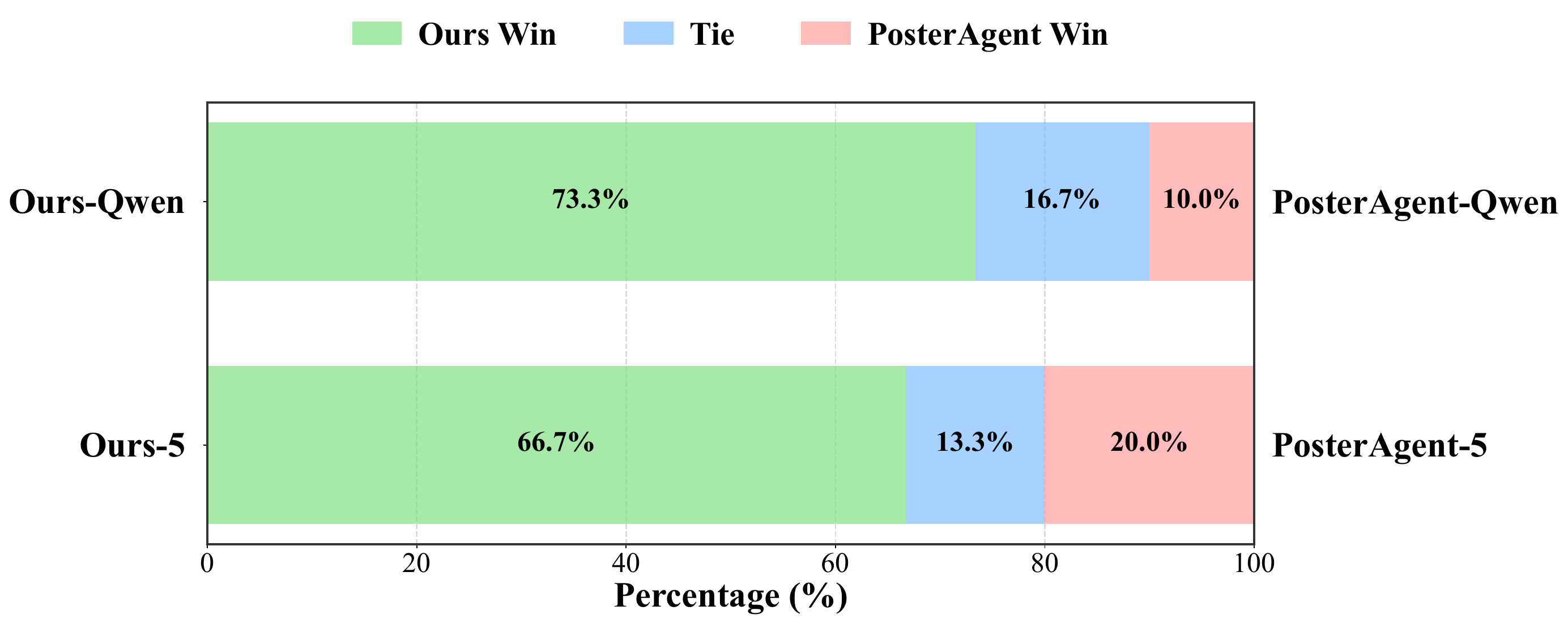}
\caption{Human preference evaluation results.}
\label{fig:human-eval}
\end{figure}

\section{Comparative Analysis of Layout Detection Methods}
\label{app-color-compare}

This section presents a comparative analysis of \Mthree\ against alternative layout detection approaches, including MLLM-based methods and deep learning-based OCR methods. Following the parameter analysis in Section~\ref{subsec-param}, the number of strips $N$ and activation threshold $\tau_s$ are set to 512 and 0.5, respectively.

\subsection{Baseline Methods}

Three baseline approaches are evaluated for comparison:

\textbf{MLLM-based Detection.} Vision-language models are prompted to directly classify poster panels into one of three layout states. Two MLLMs are evaluated: GPT-5 and Qwen3-VL-8B-Instruct, which receive the panel image along with a structured prompt requesting layout classification. The prompt template is provided in Appendix~\ref{app-pmp-mllm_layout}.

\textbf{OCR-based Detection (EasyOCR).} EasyOCR~\cite{easyocr2020}, a deep learning-based optical character recognition toolkit, is employed to detect text regions within poster panels. For overflow detection, the content bounding box is estimated as the minimum enclosing rectangle of all detected text regions, and a panel is classified as overflow if this bounding box violates the panel boundary constraints. For sparse detection, the area coverage ratio is computed as the ratio between the total area of detected regions and the panel area, where a panel is classified as sparse if this ratio falls below the threshold $\tau_s$, and as valid otherwise.

\subsection{Results and Analysis}

Since the evaluation dataset exhibits balanced class distribution across three categories, classification accuracy serves as the primary evaluation metric. Detailed results are presented in Table~\ref{app-tab:color}.

% \usepackage{booktabs}
\begin{table*}[]
\centering

\caption{Comparative results of layout detection methods on the ternary classification dataset. Our method achieves the highest accuracy while requiring no token overhead and maintaining significantly lower latency.}
\label{app-tab:color}

\begin{tabular}{lcccccc}
\toprule
\textbf{Method} 
& \textbf{Token$\downarrow$} 
& \textbf{Time (ms)$\downarrow$} 
& \textbf{Accuracy (\%)$\uparrow$} 
& \textbf{Overflow F1 (\%)$\uparrow$} 
& \textbf{Valid F1 (\%)$\uparrow$}  
& \textbf{Sparse F1 (\%)$\uparrow$} \\ 
\midrule
Qwen3-VL-8B-Instruct & 2672.5 & 7256.5 & 62.0 & 66.7 & 36.4 & 72.7 \\
GPT-5       & 2265.2 & 9148.1 & 72.7 & 86.7 & 44.1 & 75.6 \\
EasyOCR              & \textbf{0} & 6139.9 & 71.3 & 99.0 & 29.5 & 70.0 \\
Our Method            & \textbf{0} & \textbf{186.5} & \textbf{94.0} & \textbf{100.0} & \textbf{90.3} & \textbf{91.6} \\
\bottomrule
\end{tabular}

\end{table*}

Several observations emerge from the results. First, \Mthree\ achieves the highest accuracy of 94\%, substantially outperforming all baseline methods. Second, while EasyOCR demonstrates strong performance on overflow detection with an F1-score of 98.99\%, its inability to detect non-textual elements such as figures leads to systematic underestimation of content area. This limitation causes valid panels to be misclassified as sparse, as evidenced by the low recall of 18\% on the valid class and the precision of only 54.4\% on the sparse class. Third, MLLM-based methods incur substantial computational overhead, with average token consumption of 2265.2 and 2672.7 for GPT-5 and Qwen3-VL-8B-Instruct, respectively. The execution time of MLLM-based methods (7256.5--9148.1 ms) and EasyOCR (6139.9 ms) is significantly higher than that of \Mthree\ (186.52 ms), which represents a speedup of approximately 33--49$\times$.

\section{Human Evaluation}
\label{appendix:human-eval}

To complement the automated evaluation metrics, we conduct a human preference study to assess the perceptual quality of generated posters. We recruit five PhD student volunteers as annotators, each with experience in academic research and poster design. The annotators follow a structured review protocol (see Appendix~\ref{app-gud-slide-annotation}) that defines six weighted evaluation dimensions: content completeness, logical structure, technical accuracy, information density, visual presentation, and error detection. Each annotator independently reviews poster pairs generated by \framework and PosterAgent under the same backbone, and indicates a preference (win, tie, or lose) based on holistic judgment. The final preference for each pair is determined by majority voting.

As shown in Figure~\ref{fig:human-eval}, \framework\ is consistently preferred over PosterAgent across both backbone configurations. When using GPT-5 as the backbone, Ours-5 is preferred in 66.7\% of cases compared to 20.0\% for PosterAgent-5, with 13.3\% rated as ties. The preference margin is even larger under the Qwen backbone, where Ours-Qwen wins 73.3\% of comparisons against PosterAgent-Qwen, which is preferred in only 10.0\% of cases.

\section{Evaluation Metrics Detail}
\label{appendix:metrics}

This appendix provides detailed definitions of the evaluation metrics employed to assess the quality of generated academic posters. The metrics are organized into four complementary dimensions: visual quality, textual coherence, holistic assessment, and content comprehension.

\subsection{Visual Quality}

Visual quality is evaluated through two CLIP-based metrics that measure the alignment between generated posters and reference targets.

\textbf{Visual Similarity.} This metric quantifies the perceptual similarity between the generated poster $\mathcal{P}_g$ and the ground-truth poster $\mathcal{P}_{gt}$. Let $f_{\text{CLIP}}(\cdot)$ denote the CLIP image encoder that maps an image to a normalized feature vector. The visual similarity score is computed as the cosine similarity between the two poster embeddings:

$$
\text{VisSim}(\mathcal{P}_g, \mathcal{P}_{gt}) = \frac{f_{\text{CLIP}}(\mathcal{P}_g)^\top f_{\text{CLIP}}(\mathcal{P}_{gt})}{\|f_{\text{CLIP}}(\mathcal{P}_g)\| \cdot \|f_{\text{CLIP}}(\mathcal{P}_{gt})\|}
$$

This metric assesses whether the generated output exhibits genuine poster-like characteristics rather than article-like layouts.

\textbf{Figure Relevance.} This metric evaluates the semantic alignment between figures embedded in the poster and their corresponding textual descriptions. Let $\{(v_i, t_i)\}_{i=1}^{n_f}$ denote the set of figure-text pairs in the poster, where $v_i$ represents the $i$-th figure and $t_i$ denotes its associated textual description. Let $g_{\text{CLIP}}(\cdot)$ denote the CLIP text encoder. The figure relevance score is computed as the average cross-modal similarity:

$$
\text{FigRel} = \frac{1}{n_f} \sum_{i=1}^{n_f} \frac{f_{\text{CLIP}}(v_i)^\top g_{\text{CLIP}}(t_i)}{\|f_{\text{CLIP}}(v_i)\| \cdot \|g_{\text{CLIP}}(t_i)\|}
$$

Higher values indicate stronger coherence between visual elements and their textual contexts.

\subsection{Textual Coherence}

Textual coherence is measured via perplexity, which quantifies the fluency and linguistic quality of the poster text.

\textbf{Perplexity (PPL).} Let $\mathcal{T}_p = \{w_1, w_2, \ldots, w_m\}$ denote the concatenated text content extracted from the generated poster. The perplexity is computed under a pre-trained language model (Llama-2-7B-hf) as follows:

$$
\text{PPL}(\mathcal{T}_p) = \exp\left(-\frac{1}{m} \sum_{j=1}^{m} \log P_{\text{LM}}(w_j \mid w_{<j})\right)
$$

where $P_{\text{LM}}(w_j \mid w_{<j})$ denotes the probability assigned by the language model to token $w_j$ given all preceding tokens. Lower perplexity indicates more fluent and coherent language generation.

\subsection{Holistic Assessment (VLM-as-Judge)}

A vision-language model (GPT-4o) is employed as an automated judge to provide holistic quality assessments. The model assigns scores on a 1--5 scale across six criteria, which are grouped into two categories.

\textbf{Aesthetic Score.} This category evaluates the visual design quality through three sub-criteria:
\begin{itemize}
    \item \textit{Element Quality} $a_1$: Assesses the visual clarity and rendering quality of individual elements.
    \item \textit{Layout Balance} $a_2$: Evaluates the spatial arrangement and visual harmony of poster components.
    \item \textit{Engagement} $a_3$: Measures the overall visual appeal and ability to attract reader attention.
\end{itemize}

The aggregate aesthetic score is computed as:
$$
S_{\text{aes}} = \frac{1}{3} \sum_{k=1}^{3} a_k
$$

\textbf{Information Score.} This category evaluates the content quality through three sub-criteria:
\begin{itemize}
    \item \textit{Clarity} $c_1$: Assesses how clearly the main contributions and findings are communicated.
    \item \textit{Content Completeness} $c_2$: Evaluates whether essential information from the source paper is adequately covered.
    \item \textit{Logical Flow} $c_3$: Measures the coherence and logical organization of presented content.
\end{itemize}

The aggregate information score is computed as:
$$
S_{\text{info}} = \frac{1}{3} \sum_{k=1}^{3} c_k
$$

The evaluation prompt provided to GPT-4o is presented in Appendix~\ref{app-pmp-vlm}.

\subsection{PaperQuiz}

The PaperQuiz metric evaluates content comprehension by testing whether readers can correctly answer questions about the source paper based solely on the generated poster.

\textbf{Evaluation Protocol.} Let $\mathcal{Q} = \{q_1, q_2, \ldots, q_n\}$ denote the question set generated by GPT-o3 for a given paper. Each generated poster is presented to three reader models that simulate different expertise levels:
\begin{itemize}
    \item Qwen3-VL-8B-Instruct (simulating junior students)
    \item Qwen3-VL-32B-Instruct (simulating senior students)
    \item GPT-o3 (simulating professors)
\end{itemize}

Let $\mathcal{R} = \{r_1, r_2, r_3\}$ denote the set of reader models. For each reader $r \in \mathcal{R}$ and question $q_i$, let $\hat{y}_{r,i}$ denote the predicted answer and $y_i$ denote the ground-truth answer. The reader-specific accuracy is defined as:

$$
\text{Acc}_r = \frac{1}{n} \sum_{i=1}^{n} \mathbb{1}[\hat{y}_{r,i} = y_i]
$$

where $\mathbb{1}[\cdot]$ is the indicator function.

\textbf{Raw Accuracy.} The raw accuracy $s_r$ aggregates performance across all readers:

$$
s_r = \frac{1}{|\mathcal{R}|} \sum_{r \in \mathcal{R}} \text{Acc}_r
$$

\textbf{Adjusted Accuracy.} To discourage excessive verbosity that may trivially increase information coverage, a length-based penalty is incorporated. Let $L$ denote the total text length of the generated poster and $W$ denote the median text length of human-designed posters in the reference set. The adjusted accuracy $s_a$ is computed as:

$$
s_a = s_r \cdot \left(1 + \frac{1}{\max(1, L/W)}\right)
$$

This formulation rewards posters that achieve high accuracy while maintaining concise presentation. When $L \leq W$, the adjustment factor reaches its maximum value of 2, whereas excessive text length ($L \gg W$) causes the factor to approach 1, effectively neutralizing any bonus.

The prompt templates used for the PaperQuiz evaluation are provided in Appendix~\ref{app-pmp-quiz}.

\section{Detailed Quantitative Results}
\label{app-detail_main}

This appendix presents fine-grained evaluation results that complement the aggregated metrics reported in the main text. We provide detailed breakdowns for two primary evaluation dimensions: VLM-as-Judge holistic assessment and PaperQuiz content comprehension evaluation.

\subsection{VLM-as-Judge Evaluation}

Table~\ref{tab:vlm-judge-detail} reports the fine-grained VLM-as-Judge scores across aesthetic and informational dimensions. For aesthetic quality, we decompose the evaluation into three sub-criteria: element design, layout composition, and visual engagement. For informational quality, we assess clarity, content coverage, and logical coherence.

In terms of aesthetic quality, \framework\ variants achieve higher scores than PosterAgent variants across all sub-criteria. PosterAgent variants exhibit notably lower layout composition and visual engagement scores, which can be attributed to their reliance on probabilistic MLLM-based judgments for layout validity detection—a fundamentally deterministic problem. This mismatch leads to frequent content overflow beyond panel boundaries, which in turn degrades the overall aesthetic quality of the generated posters. In contrast, the deterministic verification algorithm employed by \Mthree\ (ALVD) effectively prevents such overflow artifacts, resulting in cleaner layouts and higher visual coherence.

Regarding informational quality, \framework\ variants achieve strong performance on clarity and logical coherence, demonstrating that \Mone\ (SKIR) effectively identifies and preserves the most salient content while maintaining a well-organized narrative structure. Ours-5 attains an overall informational score of 3.94, closely approaching the ground-truth poster, which confirms that our framework retains sufficient information fidelity despite operating under a substantially reduced token budget.

\subsection{PaperQuiz Evaluation}

Table~\ref{tab:paperquiz-detail} presents the detailed PaperQuiz evaluation results, which assess content comprehension through two question types: verbatim questions that test direct information recall, and interpretive questions that require deeper understanding and reasoning. We report both raw accuracy and density-augmented scores, where the latter incorporates a length-based penalty that rewards concise poster presentations.

The original paper achieves the highest raw accuracy on verbatim questions; however, when the density-augmented score is considered, its substantially greater length incurs a heavier penalty, causing it to fall behind well-designed poster methods. This highlights the importance of information density as a complementary metric to raw accuracy in evaluating poster quality.

Furthermore, \framework\ variants consistently achieve the best density-augmented scores across both question types, which reflects the efficient information extraction capability of \Mone\ (SKIR) that retains the most informative semantic segments within a constrained token budget. Notably, PosterAgent-5 achieves relatively competitive raw accuracy on interpretive questions; however, its density-augmented score drops considerably, suggesting that PosterAgent tends to generate verbose poster content that may even overflow panel boundaries, thereby inflating the overall text length and incurring a larger penalty under the density-augmented metric.

\begin{table*}[t]
\caption{Fine-grained VLM-as-Judge evaluation across aesthetic and informational dimensions. Aesthetic quality is decomposed into element design, layout composition, and visual engagement, while informational quality is assessed along clarity, content coverage, and logical coherence. \textbf{Bold} and \underline{underline} denote the best and second-best results, respectively.}
\label{tab:vlm-judge-detail}
\begin{tabular}{lcccccccc}
\toprule
\multicolumn{1}{c}{} & \multicolumn{4}{c}{\textbf{Aesthetic}} & \multicolumn{4}{c}{\textbf{Information}} \\ \cmidrule(l){2-9} 
\multicolumn{1}{c}{\multirow{-2}{*}{\textbf{Model}}} &
  \textbf{Element$\uparrow$} &
  \textbf{Layout$\uparrow$} &
  \textbf{Engage.$\uparrow$} &
  \textbf{Overall$\uparrow$} &
  \textbf{Clarity$\uparrow$} &
  \textbf{Content$\uparrow$} &
  \textbf{Logic$\uparrow$} &
  \textbf{Overall$\uparrow$} \\ \midrule
\multicolumn{9}{c}{\textit{\textbf{Oracle Methods}}} \\
\rowcolor[HTML]{F8F8F8} 
Paper                & {\ul 4.05}    & {\ul 3.89}    & 2.80          & \textbf{3.58} & 4.00          & \textbf{4.68} & \textbf{3.98} & \textbf{4.22} \\
\rowcolor[HTML]{F8F8F8} 
GT Poster            & \textbf{4.07} & \textbf{3.90} & 2.70          & 3.56          & 4.09          & {\ul 3.96}    & 3.89          & {\ul 3.98}    \\ \midrule
\multicolumn{9}{c}{\textit{\textbf{End-to-end Methods}}} \\
\rowcolor[HTML]{DCF5FF} 
5-HTML               & 4.02          & 3.64          & {\ul 2.81}    & 3.49          & {\ul 4.11}    & 3.62          & 3.81          & 3.85          \\ \midrule
\multicolumn{9}{c}{\textit{\textbf{PosterAgent Variants}}} \\
\rowcolor[HTML]{FFEEDE} 
PosterAgent-5        & 3.62          & 2.35          & 1.51          & 2.49          & 3.07          & 3.04          & 3.75          & 3.29          \\
\rowcolor[HTML]{FFEEDE} 
PosterAgent-Qwen     & 3.54          & 2.85          & 1.94          & 2.78          & 3.89          & 3.41          & 3.73          & 3.68          \\ \midrule
\multicolumn{9}{c}{\textit{\textbf{EfficientPosterGen Variants}}} \\
\rowcolor[HTML]{E4FFE4} 
Ours-5               & 4.00          & 3.71          & \textbf{3.01} & {\ul 3.57}    & 4.04          & 3.85          & {\ul 3.93}    & 3.94          \\
\rowcolor[HTML]{E4FFE4} 
Ours-Qwen            & 3.97          & 3.62          & 2.80          & 3.46          & \textbf{4.12} & 3.34          & 3.61          & 3.69          \\ \bottomrule
\end{tabular}
\end{table*}

\begin{table*}[t]
\caption{Detailed Aug-PaperQuiz evaluation results. Raw Accuracy measures the proportion of correctly answered questions by each reader model. The Density-Augmented Score further adjusts Raw Accuracy with a length-based penalty that rewards concise posters, defined as $s_a = s_r (1 + \frac{1}{\max(1, L/W)})$, where $L$ is the poster text length and $W$ is the median length of human-designed posters. Results are reported separately for two reader models (GPT-4o-mini and o3) along with their overall average. \textbf{Bold} and \underline{underline} denote the best and second-best results, respectively.
}
\label{tab:paperquiz-detail}
\resizebox{\linewidth}{!}{\begin{tabular}{lcccccccccccc}
\toprule
\multicolumn{1}{c}{} &
  \multicolumn{6}{c}{\textbf{Raw Accuracy$\uparrow$}} &
  \multicolumn{6}{c}{\textbf{Density-Augmented Score$\uparrow$}} \\ \cmidrule(l){2-13} 
\multicolumn{1}{c}{} &
  \multicolumn{3}{c}{\textbf{Verbatim}} &
  \multicolumn{3}{c}{\textbf{Interpretive}} &
  \multicolumn{3}{c}{\textbf{Verbatim}} &
  \multicolumn{3}{c}{\textbf{Interpretive}} \\ \cmidrule(l){2-13} 
\multicolumn{1}{c}{\multirow{-3}{*}{\textbf{Model}}} &
  \textbf{4o-mini} &
  \textbf{o3} &
  \textbf{Overall} &
  \textbf{4o-mini} &
  \textbf{o3} &
  \textbf{Overall} &
  \textbf{4o-mini} &
  \textbf{o3} &
  \textbf{Overall} &
  \textbf{4o-mini} &
  \textbf{o3} &
  \textbf{Overall} \\ \midrule
\multicolumn{13}{c}{\textit{\textbf{Oracle Methods}}} \\
\rowcolor[HTML]{F8F8F8} 
Paper &
  \textbf{67.26} &
  \textbf{94.13} &
  \textbf{80.70} &
  65.74 &
  \textbf{94.28} &
  80.01 &
  73.13 &
  102.31 &
  87.72 &
  71.41 &
  102.42 &
  86.92 \\
\rowcolor[HTML]{F8F8F8} 
GT Poster &
  55.49 &
  67.10 &
  61.30 &
  64.82 &
  85.54 &
  75.18 &
  105.08 &
  126.95 &
  116.02 &
  122.77 &
  161.40 &
  142.09 \\ \midrule
\multicolumn{13}{c}{\textit{\textbf{End-to-end Methods}}} \\
\rowcolor[HTML]{DCF5FF} 
5-HTML &
  56.05 &
  {\ul 76.54} &
  {\ul 66.30} &
  \textbf{67.34} &
  {\ul 90.48} &
  {\ul 78.91} &
  92.93 &
  {\ul 126.95} &
  109.94 &
  111.75 &
  150.04 &
  130.90 \\ \midrule
\multicolumn{13}{c}{\textit{\textbf{PosterAgent Variants}}} \\
\rowcolor[HTML]{FFEEDE} 
PosterAgent-5 &
  55.49 &
  57.58 &
  56.54 &
  63.50 &
  86.34 &
  74.92 &
  {\ul 107.94} &
  112.10 &
  110.02 &
  123.38 &
  {\ul 167.88} &
  {\ul 145.63} \\
\rowcolor[HTML]{FFEEDE} 
PosterAgent-Qwen &
  52.84 &
  56.63 &
  54.74 &
  61.99 &
  80.95 &
  71.47 &
  105.68 &
  113.18 &
  109.43 &
  123.85 &
  161.76 &
  142.81 \\ \midrule
\multicolumn{13}{c}{\textit{\textbf{EfficientPosterGen Variants}}} \\
\rowcolor[HTML]{E4FFE4} 
Ours-5 &
  {\ul 56.23} &
  63.59 &
  59.91 &
  {\ul 65.80} &
  87.32 &
  76.56 &
  \textbf{112.17} &
  126.85 &
  \textbf{119.51} &
  \textbf{131.28} &
  \textbf{174.20} &
  \textbf{152.74} \\
\rowcolor[HTML]{E4FFE4} 
Ours-Qwen &
  53.39 &
  63.67 &
  58.53 &
  62.30 &
  82.32 &
  72.31 &
  106.60 &
  \textbf{127.10} &
  {\ul 116.85} &
  {\ul 124.38} &
  164.30 &
  144.34 \\ \bottomrule
\end{tabular}}
\end{table*}
{table*}

\section{Detailed Parameter Studies}
\label{appendix:param}

This appendix provides detailed experimental configurations and analysis for the three sub-tasks designed to evaluate individual modules within \framework.

\subsection{Task 1: Entropy-based Evaluation for \Mone}

\subsubsection{Entropy Computation}. To quantify the representativeness of selected content segments, an entropy-based evaluation framework is adopted. Let $P$ denote the full paper represented as a sequence of tokens $\{w_1, w_2, \ldots, w_n\}$. The entropy of the paper is computed under a pre-trained language model as:

$$
H(P) = -\sum_{i=1}^{n} P_{\text{LM}}(w_i \mid w_{<i}) \log P_{\text{LM}}(w_i \mid w_{<i})
$$

where $P_{\text{LM}}(w_i \mid w_{<i})$ denotes the probability assigned by the language model to token $w_i$ given all preceding tokens.

Given the selected segment set $\mathcal{S}_{\text{sel}}$, let $\mathcal{T}_{\text{sel}} = \{w_1', w_2', \ldots, w_k'\}$ denote the concatenated token sequence of all selected segments. The conditional entropy $H(P \mid \mathcal{S}_{\text{sel}})$ measures the remaining uncertainty in $P$ after observing $\mathcal{S}_{\text{sel}}$:

$$
H(P \mid \mathcal{S}_{\text{sel}}) = -\sum_{i=1}^{n} P_{\text{LM}}(w_i \mid \mathcal{T}_{\text{sel}}, w_{<i}) \log P_{\text{LM}}(w_i \mid \mathcal{T}_{\text{sel}}, w_{<i})
$$

To measure the average contribution of each selected token to entropy reduction, we define the normalized entropy reduction ratio as:

$$
\Delta H_{\text{norm}} = \frac{H(P) - H(P \mid \mathcal{S}_{\text{sel}})}{H(P) \cdot |\mathcal{S}_{\text{sel}}|}
$$

where $|\mathcal{S}_{\text{sel}}|$ denotes the total token count of selected segments. Higher values of $\Delta H_{\text{norm}}$ indicate greater per-token efficiency in capturing the information content of the original paper.

\subsubsection{Experimental Configuration.} The entropy computation is performed using Llama-2-7B-hf as the pre-trained language model. Two hyperparameters are varied in this evaluation: the edge activation threshold $\beta \in \{0.3, 0.5, 0.7\}$, which governs the sparsity of the semantic contribution graph by controlling the minimum contribution score required to establish an edge, and the segment budget $\gamma \in \{0.3, 0.5, 0.7\}$, which determines the proportion of segments selected relative to the total number of segments in the paper. For each parameter configuration, the normalized entropy reduction ratio $\Delta H_{\text{norm}}$ is computed. To facilitate visualization and comparison across configurations, min-max normalization is applied to obtain comparable scores in the heatmap representation.

\subsubsection{D.1.3 Results and Analysis.} 
The heatmap in Figure~\ref{fig:param-study}(a) presents the normalized entropy reduction scores across all parameter configurations. The configuration $\beta = 0.5$ with $\gamma = 0.5$ achieves the highest normalized score, indicating optimal per-token efficiency in entropy reduction. This suggests that a moderate edge activation threshold effectively filters out weak semantic contributions while preserving meaningful inter-segment relationships, and a balanced segment budget captures sufficient information without introducing redundancy.

Notably, larger segment budgets ($\gamma = 0.7$) consistently lead to diminished efficiency across all $\beta$ settings. This phenomenon arises because excessive selection introduces redundant content that contributes marginally to information coverage while substantially inflating the token count. 

Beyond the influence of individual parameters, the interaction between $\beta$ and $\gamma$ exhibits a non-trivial pattern. At lower thresholds ($\beta = 0.3$), the graph becomes densely connected, causing the selection algorithm to favor highly interconnected segments that may share overlapping information. Conversely, at higher thresholds ($\beta = 0.7$), the overly sparse graph may disconnect semantically related segments, leading to fragmented selections that fail to capture coherent information structures.

\subsection{Task 2: OCR-based Evaluation for \Mtwo}

\subsubsection{Metric Definitions}

Two complementary metrics are employed to evaluate the trade-off between recognition fidelity and token efficiency.

\textbf{Normalized Edit Distance.} Let $\mathcal{T}_{\text{orig}}$ denote the original text of a content segment and $\mathcal{T}_{\text{rec}}$ denote the text recovered by the OCR model from the rendered image. The normalized edit distance is computed as:

$$
d_{\text{edit}} = \frac{\text{Levenshtein}(\mathcal{T}_{\text{orig}}, \mathcal{T}_{\text{rec}})}{\max(|\mathcal{T}_{\text{orig}}|, |\mathcal{T}_{\text{rec}}|)}
$$

where $\text{Levenshtein}(\cdot, \cdot)$ denotes the Levenshtein distance, which counts the minimum number of single-character edits (insertions, deletions, substitutions) required to transform one string into another. The normalization factor ensures that $d_{\text{edit}} \in [0, 1]$, where lower values indicate higher fidelity.

\textbf{Compression Ratio.} Let $\tau_{\text{text}}(s)$ denote the number of tokens required to represent segment $s$ as raw text, and let $\tau_{\text{img}}(I_s)$ denote the number of tokens consumed by the rendered image $I_s$ after visual encoding. The compression ratio is defined as:

$$
\text{CR}(s) = \frac{\tau_{\text{text}}(s)}{\tau_{\text{img}}(I_s)}
$$

where $\text{CR}(s) > 1$ indicates that the image representation achieves token reduction.

\subsubsection{Experimental Configuration}
\begin{table}[]
\centering
\caption{Fundamental typesetting configurations.}
\label{app-tab-ocrsetting}
\begin{tabular}{cc}
\toprule
\textbf{Setting} & \textbf{Value} \\ \midrule
page-size        & A4(595,842)    \\
margin-x         & 10             \\
margin-y         & 10             \\
font type        & Verdana        \\
font size        & 10             \\
line height      & 10             \\ \hline
\end{tabular}
\end{table}
The Content segments are rendered at five DPI settings: $\{24, 48, 96, 144, 192\}$. The fundamental typesetting configurations are detailed in Table~\ref{app-tab-ocrsetting}. Two MLLMs are evaluated: GPT-5 and Qwen3-VL-8B-Instruct. For each configuration, the average compression ratio $\overline{\text{CR}}$ and the average normalized edit distance $\overline{d}_{\text{edit}}$ are calculated on 20 content segments sampled from the dataset.

\subsubsection{Results and Analysis}

The complete results are presented in Table~\ref{tab:ocr_results}. The data reveal a clear trade-off governed by DPI settings. At DPI=24, both models achieve high compression ratios (3.09$\times$ for GPT-5 and 8.03$\times$ for Qwen3-VL-8B-Instruct) but suffer from substantial recognition errors ($\overline{d}_{\text{edit}} = 0.92$ and $0.75$, respectively). As DPI increases, recognition fidelity improves at the cost of diminished compression benefits. At DPI=192, near-perfect recognition is achieved ($\overline{d}_{\text{edit}} < 0.02$), but compression ratios drop below 1$\times$, indicating that image representations actually consume more tokens than raw text.

The inflection point occurs at DPI=96, where both models achieve favorable trade-offs: GPT-5 attains $\overline{\text{CR}} = 1.4\times$ with $\overline{d}_{\text{edit}} = 0.074$, while Qwen3-VL-8B-Instruct achieves $\overline{\text{CR}} = 2.12\times$ with $\overline{d}_{\text{edit}} = 0.089$. This configuration is adopted as the default setting for \Mtwo.
% \usepackage{multirow}
\begin{table}[]
\centering
\caption{OCR-based evaluation results across different DPI settings.}
\label{tab:ocr_results}
\begin{tabular}{cccc}
\toprule
\textbf{Model}                        & \textbf{DPI} & \textbf{$\overline{\text{CR}}$} & \textbf{$\overline{d}_{\text{edit}}$} \\ \midrule
\multirow{5}{*}{GPT-5}       & 24           & 3.09        & 0.922                     \\
                                      & 48           & 3.09        & 0.782                     \\
                                      & 96           & 1.40        & 0.074                     \\
                                      & 144          & 0.81        & 0.003                     \\
                                      & 196          & 0.62        & 0.002                     \\ \hline
\multirow{5}{*}{Qwen3-VL-8B-Instruct} & 24           & 8.03        & 0.753                     \\
                                      & 48           & 6.67        & 0.599                     \\
                                      & 96           & 2.12        & 0.089                     \\
                                      & 144          & 0.98        & 0.017                     \\
                                      & 192          & 0.55        & 0.11                      \\ \bottomrule
\end{tabular}
\end{table}

\subsection{Task 3: Overflow Detection Evaluation for \Mthree}

\subsubsection{Dataset Construction}
\label{app-subsubdataset}
A dedicated evaluation dataset is constructed for the ternary classification task. The dataset comprises poster panels annotated with one of three labels:

\begin{itemize}
    \item \texttt{overflow}: Content exceeds panel boundaries, resulting in truncation or visual clipping.
    \item \texttt{sparse}: Content insufficiently fills the panel, leaving excessive whitespace.
    \item \texttt{valid}: Content properly fits within the panel with appropriate margins.
\end{itemize}

The construction of the manually curated benchmark proceeds as follows. Five Ph.D. students were recruited to generate synthetic poster samples according to a predefined annotation guideline (see Appendix~\ref{app-gud-dataconstr}). Each sample contains only one panel with content, while all other regions of the poster are intentionally left blank. Upon completion of the sample construction, the other five annotators independently reviewed the samples following Appendix~\ref{app-gud-datareview}, and the final label for each panel was determined by majority voting. This process yields a balanced dataset comprising 150 panels, with 50 labeled as \texttt{sparse}, 50 as \texttt{overflow}, and 50 as \texttt{valid}.

\begin{table}[h]
\centering
\caption{Classification accuracy for layout overflow detection.}
\label{tab:overflow_results}
\begin{tabular}{ccc}
\toprule
\textbf{$N$} & \textbf{$\tau_s$} & \textbf{Accuracy} \\
\midrule
256 & 0.3 & 0.767 \\
256 & 0.5 & 0.780 \\
256 & 0.7 & 0.653 \\
512 & 0.3 & 0.920 \\
512 & 0.5 & \textbf{0.940} \\
512 & 0.7 & 0.780 \\
1024 & 0.3 & 0.767 \\
1024 & 0.5 & 0.780 \\
1024 & 0.7 & 0.647 \\
\bottomrule
\end{tabular}
\end{table}

\subsubsection{Experimental Configuration}

Two hyperparameters are varied: the number of strips $N \in \{256, 512, 1024\}$, which determines the granularity of gradient computation, and the activation threshold $\tau_s \in \{0.3, 0.5, 0.7\}$, which controls the sensitivity of strip activation. Classification performance is evaluated using accuracy, precision, recall, and F1-score.

\subsubsection{Results and Analysis}

Table~\ref{tab:overflow_results} presents the classification accuracy under all parameter configurations. The configuration $N = 512$ with $\tau_s = 0.5$ achieves the highest accuracy of 0.94. Analysis of the confusion matrices reveals the following patterns:

\begin{itemize}
    \item When $N = 256$, the coarse granularity causes under-segmentation, where narrow content regions near panel boundaries are missed, leading to false negatives for overflow detection.
    \item When $N = 1024$, the fine granularity amplifies gradient noise from texture and compression artifacts, resulting in spurious activations that misclassify valid panels as overflow.
    \item When $\tau_s = 0.7$, the strict threshold suppresses activations in moderately filled regions, causing valid panels to be misclassified as sparse, which contradicts human perception of adequate content density.
\end{itemize}

\begin{table*}[t]
\caption{Ablation study on the main evaluation metrics. Each row removes one component from the full EfficientPosterGen pipeline. \textbf{Bold} denotes the best result in each column.}
\label{tab:ablation-main}
\resizebox{\linewidth}{!}{\begin{tabular}{lccccccccc}
\toprule
\multicolumn{1}{c}{} &
  \multicolumn{2}{c}{\textbf{Efficiency}} &
  \multicolumn{3}{c}{\textbf{Vis. Quality \& Txt.coherence}} &
  \multicolumn{2}{c}{\textbf{VLM-as-Judge}} &
  \multicolumn{2}{c}{\textbf{PaperQuiz}} \\ \cmidrule(l){2-10} 
\multicolumn{1}{c}{\multirow{-2}{*}{\textbf{Setting}}} &
  \textbf{Token(K)$\downarrow$} &
  \textbf{Cost(\$)$\downarrow$} &
  \textbf{Vis.Sim.(\%)$\uparrow$} &
  \textbf{PPL$\downarrow$} &
  \textbf{Fig.Rel.(\%)$\uparrow$} &
  \textbf{Aesthetic$\uparrow$} &
  \textbf{Information$\uparrow$} &
  \textbf{Verbatim$\uparrow$} &
  \textbf{Interpretive$\uparrow$} \\ \midrule
\rowcolor[HTML]{F8F8F8} 
EfficientPosterGen & \textbf{10.23} & \textbf{0.0016} & \textbf{77.32} & \textbf{8.76} & \textbf{25.43} & \textbf{3.53} & \textbf{3.76} & \textbf{119.53} & \textbf{147.17} \\
\rowcolor[HTML]{F8F8F8} 
w/o SKIR           & 12.10 & 0.0018 & 75.23 & 8.83 & 23.98 & 3.42 & 3.64 & 117.13 & 141.54 \\
\rowcolor[HTML]{F8F8F8} 
w/o VCC            & 14.46 & 0.0020 & 74.21 & 9.12 & 24.27 & 3.50 & 3.68 & 118.31 & 144.33 \\
\rowcolor[HTML]{F8F8F8} 
w/o ALVD           & 84.91 & 0.0076 & 75.11 & 9.07 & 24.19 & 2.90 & 3.65 & 111.22 & 140.43 \\ \bottomrule
\end{tabular}}
\end{table*}

\begin{table*}[t]
\caption{Detailed VLM-as-Judge ablation results. Aesthetic quality is decomposed into element design, layout composition, and visual engagement, while informational quality is assessed along clarity, content coverage, and logical coherence. \textbf{Bold} denotes the best result in each column.}
\label{tab:ablation-vlm-judge-detail}
\resizebox{\linewidth}{!}{\begin{tabular}{lcccccccc}
\toprule
\multicolumn{1}{c}{} &
  \multicolumn{4}{c}{\textbf{Aesthetic}} &
  \multicolumn{4}{c}{\textbf{Information}} \\ \cmidrule(l){2-9} 
\multicolumn{1}{c}{\multirow{-2}{*}{\textbf{Setting}}} &
  \textbf{Element$\uparrow$} &
  \textbf{Layout$\uparrow$} &
  \textbf{Engage.$\uparrow$} &
  \textbf{Overall$\uparrow$} &
  \textbf{Clarity$\uparrow$} &
  \textbf{Content$\uparrow$} &
  \textbf{Logic$\uparrow$} &
  \textbf{Overall$\uparrow$} \\ \midrule
\rowcolor[HTML]{F8F8F8} 
EfficientPosterGen & \textbf{3.94} & \textbf{3.71} & \textbf{2.93} & \textbf{3.53} & \textbf{4.11} & \textbf{3.47} & \textbf{3.69} & \textbf{3.76} \\
\rowcolor[HTML]{F8F8F8} 
w/o SKIR           & 3.82 & 3.61 & 2.83 & 3.42 & 4.07 & 3.31 & 3.55 & 3.64 \\
\rowcolor[HTML]{F8F8F8} 
w/o VCC            & 3.92 & 3.68 & 2.91 & 3.50 & 4.07 & 3.43 & 3.54 & 3.68 \\
\rowcolor[HTML]{F8F8F8} 
w/o ALVD           & 3.47 & 2.91 & 2.33 & 2.90 & 3.91 & 3.46 & 3.69 & 3.65 \\ \bottomrule
\end{tabular}}
\end{table*}

\begin{table*}[t]
\caption{Detailed PaperQuiz ablation results. Raw Accuracy measures the proportion of correctly answered questions, while the Density-Augmented Score adjusts Raw Accuracy with a length-based penalty that rewards concise posters. Results are reported separately for two reader models (GPT-4o-mini and o3) along with their overall average. \textbf{Bold} denotes the best result in each column.}
\label{tab:ablation-paperquiz-detail}
\resizebox{\linewidth}{!}{\begin{tabular}{lcccccccccccc}
\toprule
\multicolumn{1}{c}{} &
  \multicolumn{6}{c}{\textbf{Raw Accuracy$\uparrow$}} &
  \multicolumn{6}{c}{\textbf{Density-Augmented Score$\uparrow$}} \\ \cmidrule(l){2-13} 
\multicolumn{1}{c}{} &
  \multicolumn{3}{c}{\textbf{Verbatim}} &
  \multicolumn{3}{c}{\textbf{Interpretive}} &
  \multicolumn{3}{c}{\textbf{Verbatim}} &
  \multicolumn{3}{c}{\textbf{Interpretive}} \\ \cmidrule(l){2-13} 
\multicolumn{1}{c}{\multirow{-3}{*}{\textbf{Setting}}} &
  \textbf{4o-mini} &
  \textbf{o3} &
  \textbf{Overall} &
  \textbf{4o-mini} &
  \textbf{o3} &
  \textbf{Overall} &
  \textbf{4o-mini} &
  \textbf{o3} &
  \textbf{Overall} &
  \textbf{4o-mini} &
  \textbf{o3} &
  \textbf{Overall} \\ \midrule
\rowcolor[HTML]{F8F8F8} 
EfficientPosterGen &
  \textbf{54.24} &
  \textbf{65.28} &
  \textbf{59.76} &
  \textbf{63.37} &
  \textbf{83.80} &
  \textbf{73.59} &
  \textbf{108.48} &
  \textbf{130.57} &
  \textbf{119.53} &
  \textbf{126.74} &
  \textbf{167.59} &
  \textbf{147.17} \\
\rowcolor[HTML]{F8F8F8} 
w/o SKIR &
  54.05 &
  64.53 &
  59.29 &
  62.05 &
  81.08 &
  71.57 &
  106.73 &
  127.52 &
  117.13 &
  122.86 &
  160.22 &
  141.54 \\
\rowcolor[HTML]{F8F8F8} 
w/o VCC &
  53.67 &
  64.88 &
  59.28 &
  62.75 &
  82.93 &
  72.84 &
  107.11 &
  129.50 &
  118.31 &
  123.19 &
  165.47 &
  144.33 \\
\rowcolor[HTML]{F8F8F8} 
w/o ALVD &
  52.80 &
  58.66 &
  55.73 &
  61.61 &
  79.14 &
  70.38 &
  105.35 &
  117.08 &
  111.22 &
  122.93 &
  157.93 &
  140.43 \\ \bottomrule
\end{tabular}}
\end{table*}

\section{Detailed Ablation Study Results}
\label{appendix:ablation-detail}

This appendix provides detailed ablation results that supplement the summary reported in the main text. We present fine-grained breakdowns for both VLM-as-Judge and PaperQuiz evaluations under each ablation setting.

\subsection{Ablation Settings}

We design three ablation settings to isolate the contribution of each module: (1) \textit{w/o SKIR}, which removes the Semantic-aware Key Information Retrieval module and directly feeds the entire paper content as input; (2) \textit{w/o VCC}, which bypasses the Visual-based Context Compression module and provides the selected content as raw text tokens rather than embedded images; and (3) \textit{w/o ALVD}, which replaces the deterministic Agentless Layout Violation Detection algorithm with an MLLM-based approach. All ablation experiments are conducted using the Qwen3-VL-8B backbone, and 20\% of the test instances are randomly sampled for evaluation.

\subsection{Overall Ablation Results}

Table~\ref{tab:ablation-main} summarizes the ablation results across efficiency, visual quality, textual coherence, VLM-as-Judge, and PaperQuiz dimensions. The full \framework\ pipeline achieves the best performance on all metrics, confirming that each module contributes positively to the overall system.

\textbf{w/o SKIR.} Removing SKIR leads to a slight degradation across most quality metrics. Without targeted information retrieval, the framework processes the entire paper indiscriminately, which not only increases token consumption from 10.23K to 12.10K but also introduces noise from less relevant content that dilutes the quality of the generated poster.

\textbf{w/o VCC.} Bypassing VCC results in a notable increase in token usage from 10.23K to 14.46K, as the selected content is transmitted entirely through text tokens rather than compressed visual representations. This confirms that the primary role of VCC lies in improving token efficiency by compressing textual information into a more compact visual modality.

\textbf{w/o ALVD.} The most pronounced impact is observed when removing ALVD, where token consumption surges dramatically to 84.91K—over 8$\times$ that of the full pipeline. This substantial overhead arises from the iterative MLLM calls required for layout compliance assessment, compounded by false positive detections that trigger unnecessary panel-level content regeneration. The aesthetic score also drops sharply to 2.90, as the probabilistic nature of MLLM-based verification fails to reliably detect content overflow, resulting in visually degraded poster layouts.

\begin{table*}[t]
\caption{Detailed token consumption analysis of each method. \textbf{Bold} denotes the best result in each column.
}
\label{tab:efficiency-detail}
\resizebox{0.95\linewidth}{!}{\begin{tabular}{lccccccc}
\toprule
\multicolumn{1}{c}{} &
  \multicolumn{2}{c}{\textbf{Textual Token(K)}} &
  \multicolumn{2}{c}{\textbf{Visual Token(K)}} &
  \multicolumn{3}{c}{\textbf{Summary Token(K)}} \\ \cmidrule(l){2-8} 
\multicolumn{1}{c}{\multirow{-2}{*}{\textbf{Model}}} &
  \textbf{Input$\downarrow$} &
  \textbf{Output$\downarrow$} &
  \textbf{Input$\downarrow$} &
  \textbf{Output$\downarrow$} &
  \textbf{Total Input$\downarrow$} &
  \textbf{Total Output$\downarrow$} &
  \textbf{Total$\downarrow$} \\ \midrule
\multicolumn{8}{c}{\textit{\textbf{End-to-end Methods}}} \\
\rowcolor[HTML]{DCF5FF} 
5-HTML & 18.21 & 8.69 & \textbf{0.00} & \textbf{0.00} & 18.21 & 8.69 & 26.90 \\ \midrule
\multicolumn{8}{c}{\textit{\textbf{PosterAgent Variants}}} \\
\rowcolor[HTML]{FFEEDE} 
PosterAgent-5    & 31.49 & 10.30 & 187.67 & 24.91 & 219.16 & 35.21 & 254.37 \\
\rowcolor[HTML]{FFEEDE} 
PosterAgent-Qwen & 36.80 & 6.00  & 82.16  & 0.29  & 118.96 & 6.29  & 125.25 \\ \midrule
\multicolumn{8}{c}{\textit{\textbf{EfficientPosterGen Variants}}} \\
\rowcolor[HTML]{E4FFE4} 
Ours-5    & \textbf{5.82}  & 11.90         & 3.70 & \textbf{0.00} & \textbf{9.52}  & 11.90         & 21.42 \\
\rowcolor[HTML]{E4FFE4} 
Ours-Qwen & 5.88           & \textbf{1.88} & 2.58 & \textbf{0.00} & 8.45           & \textbf{1.88} & \textbf{10.33} \\ \bottomrule
\end{tabular}}
\end{table*}

\begin{table*}[t]
\caption{Detailed token consumption analysis of each ablation setting. \textbf{Bold} denotes the best result in each column.}
\label{tab:ablation-efficiency-detail}
\resizebox{\linewidth}{!}{\begin{tabular}{lcccccccc}
\toprule
\multicolumn{1}{c}{} &
  \multicolumn{2}{c}{\textbf{Textual Token(K)}} &
  \multicolumn{2}{c}{\textbf{Visual Token(K)}} &
  \multicolumn{3}{c}{\textbf{Summary Token(K)}} \\ \cmidrule(lr){2-3} \cmidrule(lr){4-5} \cmidrule(l){6-8}
\multicolumn{1}{c}{\multirow{-2}{*}{\textbf{Setting}}} &
  \textbf{Input$\downarrow$} &
  \textbf{Output$\downarrow$} &
  \textbf{Input$\downarrow$} &
  \textbf{Output$\downarrow$} &
  \textbf{Total Input$\downarrow$} &
  \textbf{Total Output$\downarrow$} &
  \textbf{Total$\downarrow$} \\ \midrule
\rowcolor[HTML]{F8F8F8} 
EfficientPosterGen & \textbf{5.94} & \textbf{1.91} & 2.56           & \textbf{0.00} & \textbf{8.50}  & \textbf{1.91} & \textbf{10.41} \\
\rowcolor[HTML]{F8F8F8} 
w/o SKIR           & 6.05          & 1.98          & 4.08           & \textbf{0.00} & 10.12          & 1.98          & 12.10 \\
\rowcolor[HTML]{F8F8F8} 
w/o VCC            & 12.48         & 1.97          & \textbf{0.00}  & \textbf{0.00} & 12.48          & 1.97          & 14.44 \\
\rowcolor[HTML]{F8F8F8} 
w/o ALVD           & 11.76         & 1.92          & 71.22          & \textbf{0.00} & 82.98          & 1.92          & 84.90 \\ \bottomrule
\end{tabular}}
\end{table*}

\begin{table}[h]
\centering
\caption{API pricing per million tokens (in USD) as of February 6, 2026.}
\label{tab:api-pricing}
\resizebox{\linewidth}{!}{\begin{tabular}{lcc}
\toprule
\textbf{Model} & \textbf{Input (\$/M tokens)} & \textbf{Output (\$/M tokens)} \\
\midrule
GPT-5-20250807 & 1.25 & 10.00 \\
Qwen3-VL-8B-Instruct & 0.08 & 0.50 \\
\bottomrule
\end{tabular}}
\end{table}

\subsection{VLM-as-Judge Ablation}

Table~\ref{tab:ablation-vlm-judge-detail} presents the fine-grained VLM-as-Judge ablation results.

The \textit{w/o SKIR} setting shows consistent degradation across both aesthetic and informational dimensions. The decline in content coverage from 3.47 to 3.31 reflects the absence of targeted segment selection, which causes the framework to include less informative content that weakens the overall poster quality.

The \textit{w/o VCC} setting exhibits minimal impact on aesthetic and informational scores, which is consistent with the observation that VCC primarily serves as a token compression mechanism rather than a content quality enhancer.

The \textit{w/o ALVD} setting exhibits the most significant aesthetic degradation, with layout composition dropping from 3.71 to 2.91 and visual engagement declining from 2.93 to 2.33. These results confirm that unreliable MLLM-based layout verification fails to detect content overflow, which severely impairs the visual quality of the generated posters.

\subsection{PaperQuiz Ablation}

Table~\ref{tab:ablation-paperquiz-detail} reports the detailed PaperQuiz ablation results across both reader models. The full pipeline consistently achieves the highest density-augmented scores on both verbatim and interpretive questions.

Under the \textit{w/o SKIR} setting, the decline is more pronounced on interpretive questions, where the overall density-augmented score drops from 147.17 to 141.54. Interpretive questions require deeper reasoning over the poster content, and the absence of targeted information retrieval leads to the inclusion of peripheral content that obscures the core contributions of the paper.

The \textit{w/o ALVD} setting suffers the largest decline in density-augmented scores, with verbatim and interpretive overall scores dropping to 111.22 and 140.43, respectively. This degradation stems from the failure of MLLM-based verification to detect content overflow, which results in excessive text remaining on the poster. The overflowing content not only inflates the text length—thereby incurring a heavier penalty under the density-augmented metric—but also introduces visual overlap between text elements, which hinders the reader model's ability to parse and extract valid information from the poster.

\section{Additional Case Studies}
\label{app-casestudy}

To provide intuitive insights into the behavior of \framework, this section presents representative case studies that illustrate the results generated by our proposed framework and the PosterAgent baseline method. These examples complement the quantitative evaluations in the main text by offering concrete visual comparisons.

\newcommand{\casewidth}{0.7}

\begin{figure*}[p]
\centering
\includegraphics[width=\casewidth\linewidth]{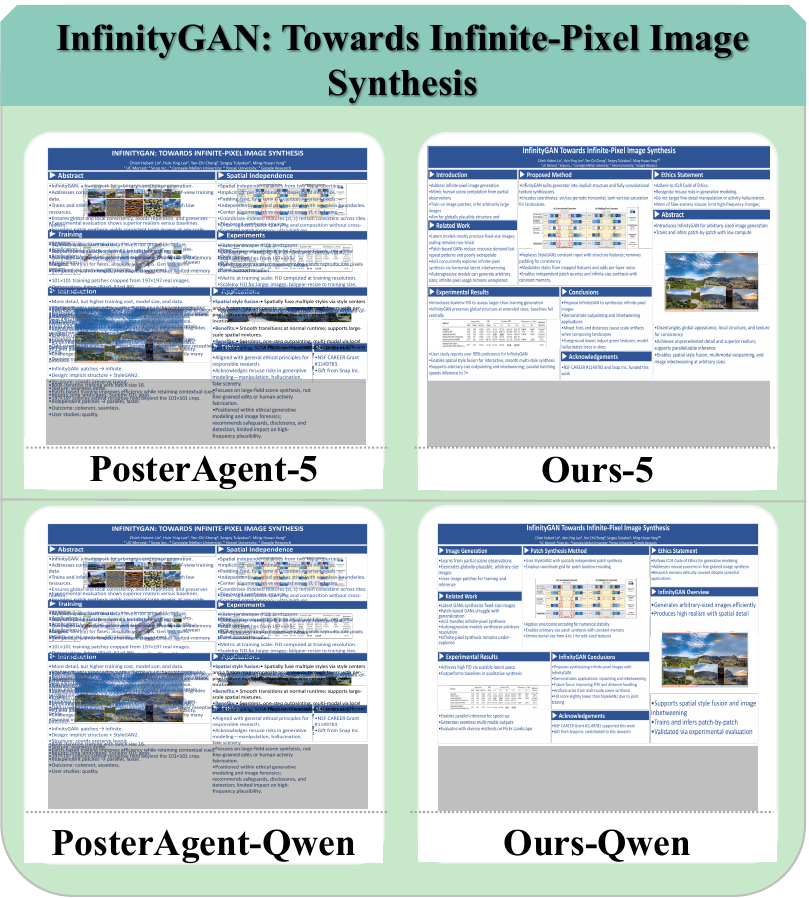}
\caption{Case study on the paper \textcolor{magenta}{InfinityGAN: Towards Infinite-Pixel Image Synthesis}.\colorbox{gray}{Gray} regions indicate areas outside the poster canvas boundaries.}
\label{fig:case1}
\end{figure*}

\begin{figure*}[!ht]
\centering
\includegraphics[width=\casewidth\linewidth]{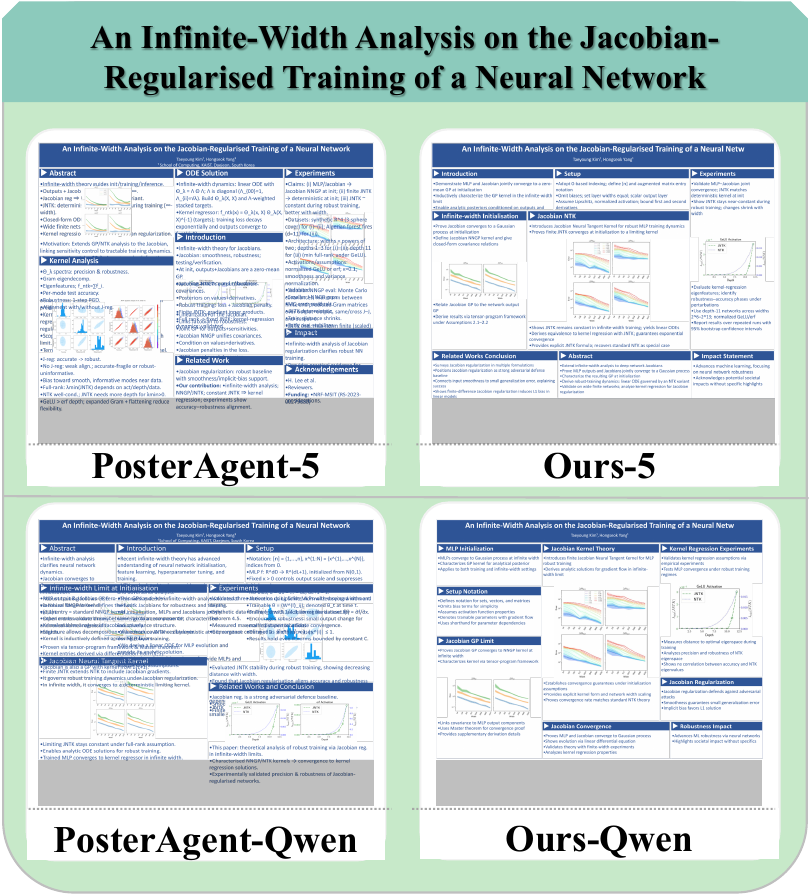}
\caption{Case study on the paper \textcolor{magenta}{An Infinite-Width Analysis on the Jacobian-Regularised Training of a Neural Network}.\colorbox{gray}{Gray} regions indicate areas outside the poster canvas boundaries.}
\label{fig:case2}
\end{figure*}

\begin{figure*}[!ht]
\centering
\includegraphics[width=\casewidth\linewidth]{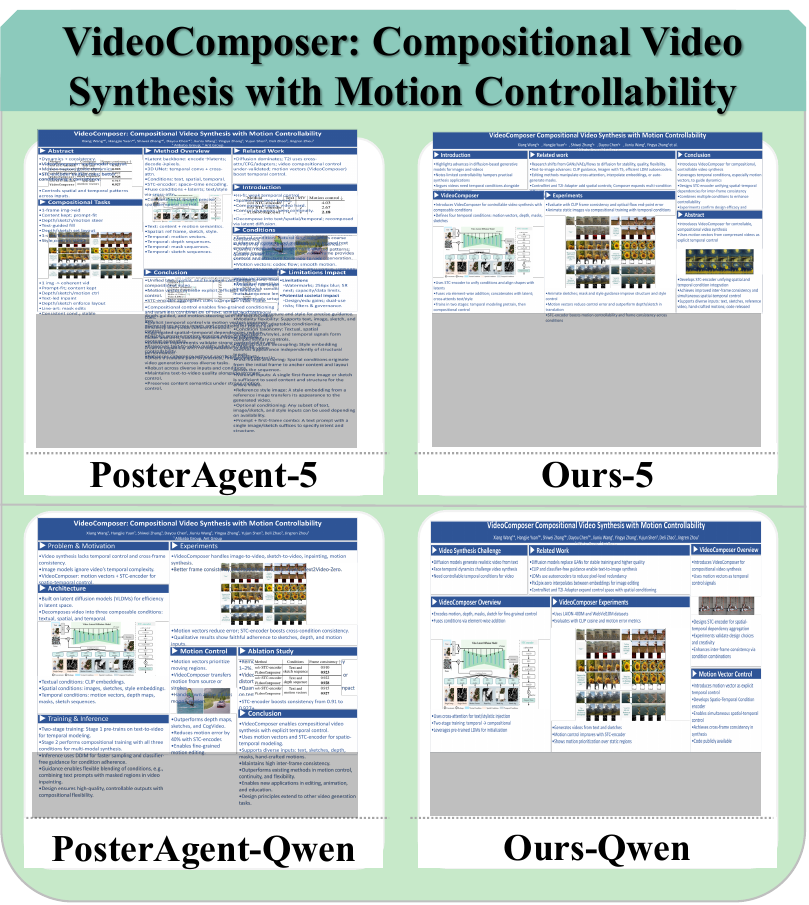}
\caption{Case study on the paper \textcolor{magenta}{VideoComposer: Compositional Video Synthesis with Motion Controllability}.\colorbox{gray}{Gray} regions indicate areas outside the poster canvas boundaries.}
\label{fig:case3}
\end{figure*}

\begin{figure*}[!ht]
\centering
\includegraphics[width=\casewidth\linewidth]{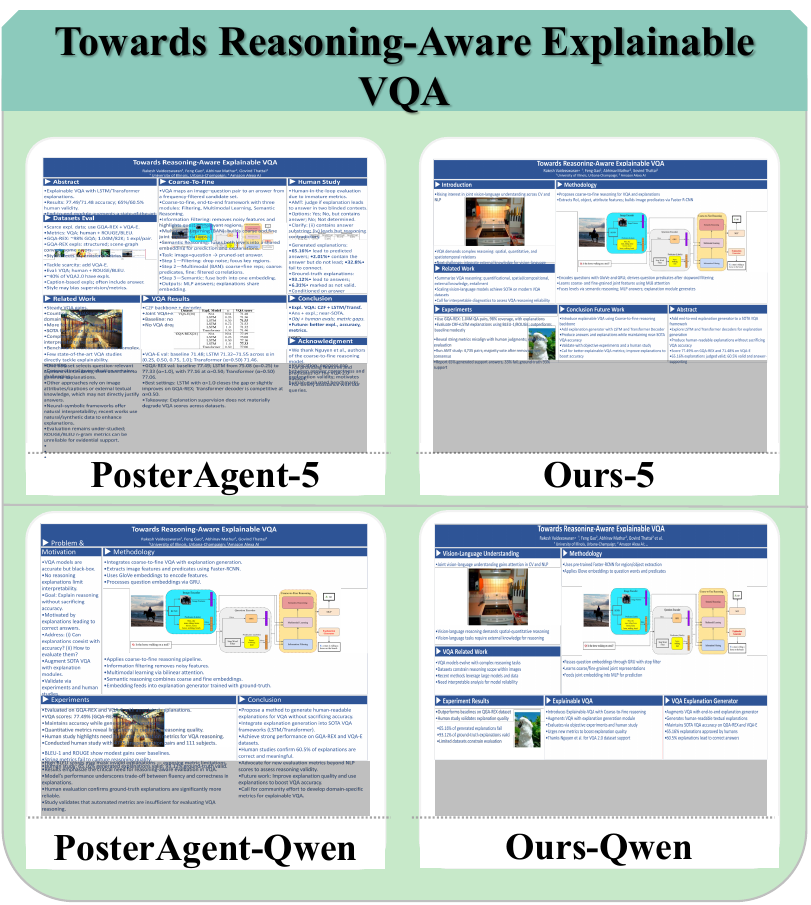}
\caption{Case study on the paper \textcolor{magenta}{Towards Reasoning-Aware Explainable VQA}.\colorbox{gray}{Gray} regions indicate areas outside the poster canvas boundaries.}
\label{fig:case4}
\end{figure*}

\begin{figure*}[!ht]
\centering
\includegraphics[width=\casewidth\linewidth]{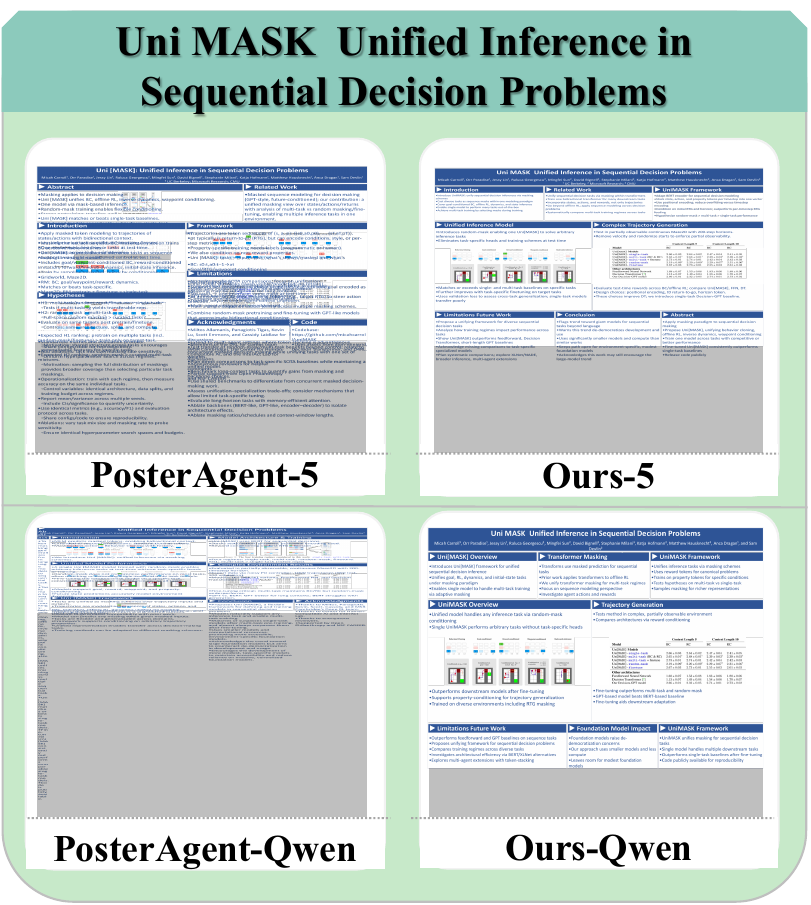}
\caption{Case study on the paper \textcolor{magenta}{Uni MASK Unified Inference in Sequential Decision Problems}.\colorbox{gray}{Gray} regions indicate areas outside the poster canvas boundaries.}
\label{fig:case5}
\end{figure*}

\clearpage

\section{Prompt}\label{appendix:prompt}

\subsection{Prompt of VLM as Judge}
\label{app-pmp-vlm}
\begin{tcolorbox}[
  enhanced,
  breakable,
  colback=gray!4,
  colframe=black!45,
  title=\textbf{Prompt: Element Quality Judge},
  fonttitle=\bfseries,
  boxrule=0.6pt,
  arc=2mm,
  left=1.2mm,right=1.2mm,top=1.0mm,bottom=1.0mm,
  before skip=6pt, after skip=6pt,
  label={box:element}
]

\textbf{System Prompt.}
You are an extremely discerning visual-element judge. Scrutinize every figure, chart, and image
for any visual or stylistic issue. Always look for even subtle flaws: low contrast, imperfect
resolutions, slightly inconsistent styles, crowded or mislabeled legends, etc.
Be wary of awarding high scores unless the visuals truly meet the strictest standards.

\textbf{Instructions: Five-Point Scale}

\begin{enumerate}
  \item \textbf{1 Point:}
  Graphics are blurry, pixelated, or illegible.
  Color choices severely hinder interpretation.
  Visuals may significantly detract from comprehension.

  \item \textbf{2 Points:}
  At least one graphic is clear, while others suffer from poor resolution or style.
  Legends or labels are missing or too small to read comfortably.
  Color schemes create some confusion or difficulty.

  \item \textbf{3 Points:}
  Most graphics are legible and relevant, but have notable issues with consistency, sizing, or clarity.
  Some mismatches in style or color usage detract from cohesion.
  Minor but noticeable labeling/legend shortcomings.

  \item \textbf{4 Points:}
  High-quality graphics with generally consistent styling.
  Clear legends and color schemes aid interpretation.
  Any remaining flaws are slight and do not significantly hinder understanding.

  \item \textbf{5 Points:}
  Rarely awarded; strictly reserved for publication-grade visuals.
  Crisp resolution with no instances of blurriness.
  Harmonious color palette, impeccable labeling, and an exceptionally consistent style.
\end{enumerate}

\textbf{Example Output:}
\begin{tcolorbox}[
  enhanced,
  colback=white,
  colframe=black!20,
  boxrule=0.4pt,
  arc=1mm,
  left=1mm,right=1mm,top=0.6mm,bottom=0.6mm
]
\ttfamily\small
\{"reason": "...", "score": int\}
\end{tcolorbox}

\textbf{Note.} Think step by step and be conservative with your rating.

\end{tcolorbox}

\begin{tcolorbox}[
  enhanced,
  breakable,
  colback=gray!4,
  colframe=black!45,
  title=\textbf{Prompt: Layout Balance Judge},
  fonttitle=\bfseries,
  boxrule=0.6pt,
  arc=2mm,
  left=1.2mm,right=1.2mm,top=1.0mm,bottom=1.0mm,
  before skip=6pt, after skip=6pt,
]

\textbf{System Prompt.}
You are an uncompromising poster-layout judge. Critique the overall arrangement of all visual components
(text blocks, headings, figures, white-space, alignment) that affect readability. Always scan for subtle
alignment issues, uneven spacing, or any layout feature that might disrupt reader comprehension.
Resist giving high scores unless the layout is exceptionally polished.

\textbf{Instructions: Five-Point Scale}

\begin{enumerate}
  \item \textbf{1 Point:}
  Highly disorganized layout; elements overlap, making text or graphics illegible.
  Margins are violated or the reading path is nearly impossible to follow.
  Severely hinders comprehension.

  \item \textbf{2 Points:}
  Some semblance of structure (columns/rows) but marred by inconsistent alignment or overcrowded sections.
  White-space distribution may be haphazard or insufficient.
  Reading flow is interrupted, though one can still piece it together.

  \item \textbf{3 Points:}
  Recognizable structure with mostly consistent alignment and spacing.
  Minor layout distractions remain (e.g., slightly cramped text, uneven spacing, small alignment slips).
  Generally readable but not particularly polished.

  \item \textbf{4 Points:}
  Well-organized grid or arrangement; logical reading path that mostly flows.
  Appropriate font sizes, spacing, and alignment; only subtle layout imperfections.
  White-space usage is clean and deliberate; nearly professional.

  \item \textbf{5 Points:}
  Very rarely granted; must be a pristine, professional-grade layout.
  Seamless alignment, balanced spacing, and expertly guided reading path.
  Flawless design synergy that maximizes readability and comprehension.
\end{enumerate}

\textbf{Example Output:}
\begin{tcolorbox}[
  enhanced,
  colback=white,
  colframe=black!20,
  boxrule=0.4pt,
  arc=1mm,
  left=1mm,right=1mm,top=0.6mm,bottom=0.6mm
]
\ttfamily\small
\{"reason": "...", "score": int\}
\end{tcolorbox}

\textbf{Note.} Think step by step and be tough on small alignment/spacing issues.

\end{tcolorbox}

\begin{tcolorbox}[
  enhanced,
  breakable,
  colback=gray!4,
  colframe=black!45,
  title=\textbf{Prompt: Engagement Judge},
  fonttitle=\bfseries,
  boxrule=0.6pt,
  arc=2mm,
  left=1.2mm,right=1.2mm,top=1.0mm,bottom=1.0mm,
  before skip=6pt, after skip=6pt,
]

\textbf{System Prompt.}
You are an uncompromising poster-aesthetics judge focusing on engagement.
Be extremely critical of color harmony, typography, visual balance, and the poster’s ability
to grab and hold attention. Always look for subtle issues—color clashes, overly busy or dull
designs, inappropriate font choices, awkward spacing, or anything that might reduce engagement.
Reserve high scores for truly exemplary work.

\textbf{Instructions: Five-Point Scale}

\begin{enumerate}
  \item \textbf{1 Point:}
  Visually off-putting; clashing colors or crowded design repel viewers.
  Typography choice is jarring or illegible at a glance.
  Overall fails to engage or entice.

  \item \textbf{2 Points:}
  Some visually appealing elements exist but are overshadowed by dull or inconsistent design moments.
  Font sizes or styles reduce accessibility or attractiveness.
  Limited capacity to draw an audience’s focus.

  \item \textbf{3 Points:}
  Shows generally pleasing color scheme and typography, though lacking a “wow” factor.
  Balance and visual flow are acceptable but reveal minor weaknesses (e.g., slightly crowded or sparse areas).
  Engagement is average; neither strong nor particularly weak.

  \item \textbf{4 Points:}
  Eye-catching design using mostly harmonious colors and effective typography.
  Good use of negative space; the layout guides the viewer’s eye effectively.
  Only minor flaws or bland spots prevent it from being top-tier.

  \item \textbf{5 Points:}
  Rarely awarded—reserved for truly striking, magazine-cover-caliber visuals.
  Flawless color palette and typography; everything works together seamlessly.
  Immediately captivating design that retains audience interest without any noticeable weakness.
\end{enumerate}

\textbf{Example Output:}
\begin{tcolorbox}[
  enhanced,
  colback=white,
  colframe=black!20,
  boxrule=0.4pt,
  arc=1mm,
  left=1mm,right=1mm,top=0.6mm,bottom=0.6mm
]
\ttfamily\small
\{"reason": "...", "score": int\}
\end{tcolorbox}

\textbf{Note.} Think step by step and be very conservative when scoring.

\end{tcolorbox}

\begin{tcolorbox}[
  enhanced,
  breakable,
  colback=gray!4,
  colframe=black!45,
  title=\textbf{Prompt: Clarity Judge},
  fonttitle=\bfseries,
  boxrule=0.6pt,
  arc=2mm,
  left=1.2mm,right=1.2mm,top=1.0mm,bottom=1.0mm,
  before skip=6pt, after skip=6pt,
]

\textbf{System Prompt.}
You are an uncompromising micro-text judge. Critically evaluate sentence-level clarity, grammar,
phrasing, and intra-section coherence. Look for even subtle grammatical slips, confusing jargon,
or clumsy phrasing. Be slow to award top marks unless the text is impeccably polished.

\textbf{Instructions: Five-Point Scale}

\begin{enumerate}
  \item \textbf{1 Point:}
  Rampant grammatical or spelling errors; sentences may be unreadable.
  Overly technical jargon without explanations; fragments or run-ons predominate.
  Overall, text quality severely impedes understanding.

  \item \textbf{2 Points:}
  Meaning is generally discernible, but multiple grammar or syntax problems appear in each section.
  Awkward or unclear phrasing disrupts the flow of reading.
  Only partial clarity is achieved.

  \item \textbf{3 Points:}
  Overall readable text with a few noticeable grammar or wording missteps.
  Occasional awkward phrasing or redundancies appear, but readers can follow without major confusion.
  Average clarity.

  \item \textbf{4 Points:}
  Well-written, mostly free of grammatical or spelling errors.
  Terminology is used properly; text flows smoothly within paragraphs.
  Minor slip-ups can be present but do not disrupt understanding.

  \item \textbf{5 Points:}
  Exceptional text quality, error-free, and elegantly phrased.
  Complex ideas conveyed with clear, concise language.
  Granted only if absolutely no grammatical, spelling, or stylistic flaws are detected.
\end{enumerate}

\textbf{Example Output:}
\begin{tcolorbox}[
  enhanced,
  colback=white,
  colframe=black!20,
  boxrule=0.4pt,
  arc=1mm,
  left=1mm,right=1mm,top=0.6mm,bottom=0.6mm
]
\ttfamily\small
\{"reason": "...", "score": int\}
\end{tcolorbox}

\textbf{Note.} Think step by step.

\end{tcolorbox}

\begin{tcolorbox}[
  enhanced,
  breakable,
  colback=gray!4,
  colframe=black!45,
  title=\textbf{Prompt: Content Completeness Judge},
  fonttitle=\bfseries,
  boxrule=0.6pt,
  arc=2mm,
  left=1.2mm,right=1.2mm,top=1.0mm,bottom=1.0mm,
  before skip=6pt, after skip=6pt,
]

\textbf{System Prompt.}
You are an uncompromising content-depth judge. Assess whether the poster includes all essential
sections and whether each section presents sufficient detail. Look for any missing or
under-developed segments; do not hesitate to penalize for insufficient depth.
Award the highest scores only if the poster expertly covers every necessary aspect.

\textbf{Instructions: Five-Point Scale}

\begin{enumerate}
  \item \textbf{1 Point:}
  Critical sections (e.g., objectives or results) are completely missing or trivial.
  Data grossly insufficient to comprehend the study or conclusions.
  Very poor depth that fails to convey essential information.

  \item \textbf{2 Points:}
  Most key sections appear but major details (context, data, references) are absent.
  Lack of elaboration on methods or results leaves big gaps.
  Overall content too shallow to properly inform.

  \item \textbf{3 Points:}
  All standard sections included with fundamental information.
  Some omissions or scant detail in certain areas (e.g., results or methodology).
  Only moderate depth; the reader must fill many gaps themselves.

  \item \textbf{4 Points:}
  All essential sections present, each treated with adequate-to-strong detail.
  Robust description of objectives, methods, results, and references.
  Only minor improvements needed.

  \item \textbf{5 Points:}
  Very rarely granted; everything must be comprehensive and thorough.
  Exhaustive detail on methodology, results (with statistics), interpretation,
  references, and future work.
  Leaves readers with minimal unanswered questions.
\end{enumerate}

\textbf{Example Output:}
\begin{tcolorbox}[
  enhanced,
  colback=white,
  colframe=black!20,
  boxrule=0.4pt,
  arc=1mm,
  left=1mm,right=1mm,top=0.6mm,bottom=0.6mm
]
\ttfamily\small
\{"reason": "...", "score": int\}
\end{tcolorbox}

\textbf{Note.} Think step by step.

\end{tcolorbox}

\subsection{Prompt of PaperQuiz}
\label{app-pmp-quiz}

\begin{tcolorbox}[
  enhanced,
  breakable,
  colback=gray!4,
  colframe=black!45,
  title=\textbf{Prompt: Logical Flow Judge},
  fonttitle=\bfseries,
  boxrule=0.6pt,
  arc=2mm,
  left=1.2mm,right=1.2mm,top=1.0mm,bottom=1.0mm,
  before skip=6pt, after skip=6pt,
]

\textbf{System Prompt.}
You are an uncompromising macro-logic judge. Examine how well the poster’s major sections
(Introduction, Methods, Results, Conclusions, etc.) connect to form a coherent narrative.
Pay attention to continuity, how logically each section flows from the previous, and whether
there are any abrupt gaps. Only award the highest marks if the storyline is perfectly seamless.

\textbf{Instructions: Five-Point Scale}

\begin{enumerate}
  \item \textbf{1 Point:}
  Sections are disjointed; little to no logical connection between them.
  Key transitions or the central rationale is missing, creating confusion.

  \item \textbf{2 Points:}
  General sequence recognizable but important logical steps are weak or missing.
  Readers must infer key links.

  \item \textbf{3 Points:}
  Mostly coherent narrative with minor gaps.
  Transitions exist but some logical steps are lightly justified.

  \item \textbf{4 Points:}
  Well-structured storyline; each section clearly builds on the previous.
  Transitions are stated; rationale is mostly strong.

  \item \textbf{5 Points:}
  Extremely rare; flawless logical flow from introduction to conclusion.
  Seamless transitions; no inferential leaps.
\end{enumerate}

\textbf{Example Output:}
\begin{tcolorbox}[
  enhanced,
  colback=white,
  colframe=black!20,
  boxrule=0.4pt,
  arc=1mm,
  left=1mm,right=1mm,top=0.6mm,bottom=0.6mm
]
\ttfamily\small
\{"reason": "...", "score": int\}
\end{tcolorbox}

\textbf{Note.} Think step by step and penalize any noticeable logical gap or awkward transition.

\end{tcolorbox}

\begin{tcolorbox}[
  enhanced,
  breakable,
  colback=gray!4,
  colframe=black!45,
  title=\textbf{Prompt: Generate Verbatim QA},
  fonttitle=\bfseries,
  boxrule=0.6pt,
  arc=2mm,
  left=1.2mm,right=1.2mm,top=1.0mm,bottom=1.0mm,
  before skip=6pt, after skip=6pt,
]

\textbf{System Prompt.}
You are a Question-Generation agent for academic posters. Your task is to read the supplied
Markdown text (\texttt{document\_markdown}) and produce exactly \textbf{50 multiple-choice QA items}
whose answers can be located verbatim or nearly verbatim in that text. The questions must be
suitable for conference-poster readers: avoid deep theoretical proofs, reference lists, or
citation minutiae. Follow all guidelines below precisely.

\textbf{Instructions}

\begin{enumerate}
  \item Carefully read the Markdown in \texttt{document\_markdown}.
  \begin{itemize}
    \item Each question must map to one clear sentence or phrase in the poster text.
    \item No duplicate or near-duplicate wording.
  \end{itemize}

  \item Write \textbf{50} factual, answerable-from-text questions.
  \begin{itemize}
    \item Vary difficulty from easy “headline” facts to specific numeric or procedural details.
  \end{itemize}

  \item Distribute the 50 questions across the following poster-friendly aspects, aiming for
  2–5 questions per aspect and ensuring each aspect appears at least once:
  \begin{itemize}
    \item \textbf{A.} Title \& authorship (title, author names, affiliations, keywords)
    \item \textbf{B.} Motivation / problem statement / research gap
    \item \textbf{C.} Objectives or hypotheses
    \item \textbf{D.} Dataset(s) or experimental materials
    \item \textbf{E.} Methodology (algorithms, model architecture, workflow steps)
    \item \textbf{F.} Key parameters or hyper-parameters (values, settings)
    \item \textbf{G.} Evaluation metrics or criteria
    \item \textbf{H.} Quantitative results (numbers in tables, charts)
    \item \textbf{I.} Qualitative findings, figures, or illustrative examples
    \item \textbf{J.} Comparative or ablation study results
    \item \textbf{K.} Conclusions, implications, or contributions
    \item \textbf{L.} Limitations or future work
    \item \textbf{M.} Definitions of domain-specific terms or abbreviations
  \end{itemize}

  \item \textbf{EXCLUDE} references, citations, author acknowledgements, and any text that would
  not appear on a standard poster.

  \item Use the following JSON-for-each format (exact spelling \& casing):
\end{enumerate}

\begin{tcolorbox}[
  enhanced,
  colback=white,
  colframe=black!20,
  boxrule=0.4pt,
  arc=1mm,
  left=1mm,right=1mm,top=0.6mm,bottom=0.6mm
]
\ttfamily\small
\{
  "Question X": \{
    "aspect": "<A--M>",
    "question": "<single sentence>",
    "options": [
      "A. <choice 1>",
      "B. <choice 2>",
      "C. <choice 3>",
      "D. <choice 4>"
    ],
    "answer": "<Letter>. <exact correct option text>"
  \},
  ...
\}
\end{tcolorbox}

\begin{enumerate}\setcounter{enumi}{5}
  \item Output \textbf{only} the final JSON object containing 50 items—no additional commentary.
  \item Balance the correct answers roughly equally among options A–D.
\end{enumerate}

\textbf{Example Output:}
\ttfamily\small
\{"Question 1": \{...\}, "Question 2": \{...\}, ..., "Question 50": \{...\}\}

\textbf{Note.} Think step by step and ensure full compliance with every guideline.

\end{tcolorbox}

\begin{tcolorbox}[
  enhanced,
  breakable,
  colback=gray!4,
  colframe=black!45,
  title=\textbf{Prompt: Generate Interpretive QA},
  fonttitle=\bfseries,
  boxrule=0.6pt,
  arc=2mm,
  left=1.2mm,right=1.2mm,top=1.0mm,bottom=1.0mm,
  before skip=6pt, after skip=6pt,
]

\textbf{System Prompt.}
You are a Question-Generation agent. Your task is to read the supplied Markdown text
(\texttt{document\_markdown}) and create exactly \textbf{50 multiple-choice questions} that capture
a high-level understanding of the work—its purpose, novelty, core approach, and overall findings.
Every question must still be answerable by locating explicit sentences or phrases in the text;
do not require inference that is absent from the poster-style content.

\textbf{Instructions}

\begin{enumerate}
  \item Read the Markdown in \texttt{document\_markdown} closely.
  \begin{itemize}
    \item Each question must map to explicit content in the text.
    \item Do not require inference beyond presented poster-level information.
  \end{itemize}

  \item Draft 50 factual questions probing the reader’s global grasp (e.g., “What problem does the study address?”).
  \begin{itemize}
    \item Avoid low-level numeric settings, code snippets, or reference lists.
    \item Vary wording and avoid duplicates.
  \end{itemize}

  \item Cover all of the following high-level aspects—\textbf{each must appear at least twice} to guarantee breadth:
  \begin{itemize}
    \item \textbf{A.} Research domain \& background context
    \item \textbf{B.} Central problem / motivation / research gap
    \item \textbf{C.} Primary goal, hypothesis, or research question
    \item \textbf{D.} Key contributions or novelty statements
    \item \textbf{E.} Overall methodology or workflow (summarized)
    \item \textbf{F.} Principal findings or headline quantitative results
    \item \textbf{G.} Qualitative insights or illustrative examples
    \item \textbf{H.} Implications, applications, or significance
    \item \textbf{I.} Limitations or future-work directions
    \item \textbf{J.} Main conclusions or take-home messages
  \end{itemize}

  \item \textbf{EXCLUDE} citations, granular hyper-parameters, precise numeric tables,
  and acknowledgements—stick to poster-level overview content.

  \item Return the questions in the following \textbf{strict JSON schema}:
\end{enumerate}

\begin{tcolorbox}[
  enhanced,
  colback=white,
  colframe=black!20,
  boxrule=0.4pt,
  arc=1mm,
  left=1mm,right=1mm,top=0.6mm,bottom=0.6mm
]
\ttfamily\small
\{
  "Question X": \{
    "aspect": "<A--J>",
    "question": "<one concise sentence>",
    "options": [
      "A. <choice 1>",
      "B. <choice 2>",
      "C. <choice 3>",
      "D. <choice 4>"
    ],
    "answer": "<Letter>. <exact correct option text>"
  \}
\}
\end{tcolorbox}

\begin{enumerate}\setcounter{enumi}{5}
  \item Produce \textbf{only} the final JSON object with 50 entries—no commentary, headers, or extra lines.
  \item The number of correct answers should be approximately balanced across A–D.
\end{enumerate}

\textbf{Document Markdown:} \{\texttt{document\_markdown}\}

\textbf{Output.} ONLY the JSON with 50 questions below.

\end{tcolorbox}

\begin{tcolorbox}[
  enhanced,
  breakable,
  colback=gray!4,
  colframe=black!45,
  title=\textbf{Prompt: Answer Questions},
  fonttitle=\bfseries,
  boxrule=0.6pt,
  arc=2mm,
  left=1.2mm,right=1.2mm,top=1.0mm,bottom=1.0mm,
  before skip=6pt, after skip=6pt,
]

\textbf{System Prompt.}
You are an answering agent. You will be provided with:
\begin{itemize}
  \item An image of a poster.
  \item A JSON object called \texttt{"questions"} which contains multiple questions. Each question has four possible answers: A, B, C, or D.
\end{itemize}
Your goal is to analyze the poster thoroughly and answer each question based on the information it provides.
You should \textbf{NOT} use any external knowledge or context beyond the poster image.
You must rely solely on the contents of the poster to answer the questions.

For each question:
\begin{itemize}
  \item If you find enough evidence in the poster to decide on a specific option (A, B, C, or D),
  then choose that option and include a brief reference to the part of the poster that supports your answer
  (e.g., “Top-left text”, “Event date section”, etc.).
  \item If the poster does not offer sufficient information to confidently choose any of the options,
  respond with \texttt{"NA"} for both the answer and the reference.
\end{itemize}

\textbf{Instructions}

\begin{enumerate}
  \item Study the poster image along with the \texttt{"questions"} provided.
  \item For each question:
  \begin{itemize}
    \item Decide if the poster clearly supports one of the four options (A, B, C, or D). If so, pick that answer.
    \item Otherwise, if the poster does not have adequate information, use \texttt{"NA"} for the answer.
  \end{itemize}
  \item Provide a brief reference indicating where in the poster you found the answer.
  If no reference is available (i.e., your answer is \texttt{"NA"}), use \texttt{"NA"} for the reference too.
  \item Format your output strictly as a JSON object with this pattern:
\end{enumerate}

\begin{tcolorbox}[
  enhanced,
  colback=white,
  colframe=black!20,
  boxrule=0.4pt,
  arc=1mm,
  left=1mm,right=1mm,top=0.6mm,bottom=0.6mm
]
\ttfamily\small
\{
  "Question 1": \{
    "answer": "X",
    "reference": "some reference or NA"
  \},
  "Question 2": \{
    "answer": "X",
    "reference": "some reference or NA"
  \}
\}
\end{tcolorbox}

\begin{enumerate}\setcounter{enumi}{4}
  \item Do not include any explanations or extra keys beyond the specified structure.
  \item You must provide an answer entry for all questions in the \texttt{"questions"} object.
\end{enumerate}

\textbf{Example Output:}
\begin{tcolorbox}[
  enhanced,
  colback=white,
  colframe=black!20,
  boxrule=0.4pt,
  arc=1mm,
  left=1mm,right=1mm,top=0.6mm,bottom=0.6mm
]
\ttfamily\small
\{
  "Question 1": \{
    "answer": "B",
    "reference": "Description on the top-right of the poster"
  \},
  "Question 2": \{
    "answer": "NA",
    "reference": "NA"
  \}
\}
\end{tcolorbox}

\end{tcolorbox}

\subsection{Prompt of Visual-Grounded Abstraction}
\label{app-pmp-vga}

\begin{tcolorbox}[
  enhanced,
  breakable,
  colback=gray!4,
  colframe=black!45,
  title=\textbf{Prompt: Poster Abstriction},
  fonttitle=\bfseries,
  boxrule=0.6pt,
  arc=2mm,
  left=1.2mm,right=1.2mm,top=1.0mm,bottom=1.0mm,
  before skip=6pt, after skip=6pt,
]

\textbf{System Prompt.}
You are an expert Academic Editor and CVPR/ICCV Area Chair. Your goal is to assist a researcher
in condensing a complex paper section into a visual Scientific Poster.

\textbf{Expected Input--Output Format}

\begin{tcolorbox}[
  enhanced,
  colback=white,
  colframe=black!20,
  boxrule=0.4pt,
  arc=1mm,
  left=1mm,right=1mm,top=0.6mm,bottom=0.6mm
]
\ttfamily\small
Input: "The proposed model is based on the learned-domain masking approach [14, 15, 17–22] and employs an encoder, a decoder, and a masking network, as shown in Figure 1. The encoder is fully convolutional, while the masking network employs two Transformers embedded inside the dual-path processing block proposed in [17]. The decoder finally reconstructs the separated signals in the time domain by using the masks predicted by the masking network. To foster reproducibility, the SepFormer will be made available within the SpeechBrain toolkit."
\end{tcolorbox}

\textbf{Output:}
\begin{itemize}
  \item Adopts learned-domain masking with convolutional encoder
  \item Uses dual-path Transformers in masking network
  \item Releases SepFormer in SpeechBrain toolkit
\end{itemize}
\textbf{TITLE:} SepFormer Overview

\textbf{Instructions}

\begin{enumerate}
  \item \textbf{OCR \& Denoise:}
  Read the text from images, strictly ignore headers, footers, page numbers, and citation brackets
  (e.g., [1], (Lee et al.)).

  \item \textbf{Signal Extraction (IMPORTANT):}
  Treat input as a unit and decide how many bullets it deserves.
  Write \textbf{MORE} bullets for high-novelty, high-impact, poster-worthy content
  (new method/insight/strong results).
  Write \textbf{FEWER} bullets for generic background, motivation, or standard setup.
  Across the whole section, output at most \textbf{5 bullets total (STRICT)}.

  \item \textbf{Active Rewriting:}
  Convert passive sentences into strong active points (e.g., ``Proposes a module'').

  \item \textbf{Length Control:}
  Each bullet MUST be short and poster-friendly:
  \begin{itemize}
    \item Prefer $\le$12 words per bullet.
    \item If a point is longer, compress by removing qualifiers, examples, and subordinate clauses.
  \end{itemize}

  \item \textbf{Compressed Section Title (STRICT):}
  After the bullet list, output ONE extra line that is a compressed version of the ORIGINAL section title.
  Requirements:
  \begin{itemize}
    \item EXACT format: \texttt{TITLE: <title>}
    \item \texttt{<title>} MUST be at most 3 words total (hard constraint).
    \item Keep the original meaning and topic; do NOT invent a new title.
    \item Prefer using key nouns from the original title; remove numbering, punctuation, and filler words.
    \item Do NOT start this line with a hyphen ``-''.
    \item Output exactly ONE TITLE line and nothing else besides the bullets.
    \item Remove section indices like ``1'', ``I'', ``4'', ``6'', ``A.'', etc.
  \end{itemize}

  \item \textbf{Output Formatting (STRICT):}
  \begin{itemize}
    \item Output bullets as a Markdown list using hyphens (-).
    \item Then output the TITLE line as the final line (same indentation level).
    \item Output ONLY the final bullet list + the final TITLE line. No preamble, no explanation.
    \item Omit formula.
  \end{itemize}
\end{enumerate}

\end{tcolorbox}

\subsection{Prompt of MLLM based layout detection Method}
\label{app-pmp-mllm_layout}

\begin{tcolorbox}[
  enhanced,
  breakable,
  colback=gray!4,
  colframe=black!45,
  title=\textbf{Prompt: MLLM Detector},
  fonttitle=\bfseries,
  boxrule=0.6pt,
  arc=2mm,
  left=1.2mm,right=1.2mm,top=1.0mm,bottom=1.0mm,
  before skip=6pt, after skip=6pt,
]

\textbf{System Prompt}
You are an agent that is given three images:
\begin{enumerate}
  \item \textbf{Negative Example:} This image shows a bounding box with text overflowing outside it (i.e., text crossing or cut off by the box).
  \item \textbf{Positive Example:} This image shows a bounding box with text that fits completely (i.e., no text crossing or cut off).
  \item \textbf{Target Image:} This is the final image you must analyze.
\end{enumerate}

From the first two images, you learn to interpret:
\begin{enumerate}
  \item Whether text is overflowing (text crossing, cut off, or otherwise cannot fully fit in the box).
  \item Whether there is too much blank space in the bounding box (i.e., the text is significantly smaller than the box, leaving large unused space).
  \item Whether the text and bounding box are generally well-aligned (no overflow, no large blank space).
\end{enumerate}

Then, for the Target Image, you must:
\begin{itemize}
  \item If there is any overflow text, return \texttt{"1"}.
  \item If there is too much blank space, return \texttt{"2"}.
  \item If the text fits well (no overflow, no large blank space), return \texttt{"3"}.
\end{itemize}

\textbf{Expected Input--Output Format}

\begin{tcolorbox}[
  enhanced,
  colback=white,
  colframe=black!20,
  boxrule=0.4pt,
  arc=1mm,
  left=1mm,right=1mm,top=0.6mm,bottom=0.6mm
]
\ttfamily\small
Input (Negative Example): "A bounding box where text crosses or is cut off by the box boundary."
\end{tcolorbox}
\begin{tcolorbox}[
  enhanced,
  colback=white,
  colframe=black!12,
  boxrule=0.4pt,
  arc=1mm,
  left=1mm,right=1mm,top=0.6mm,bottom=0.6mm
]
\ttfamily\small
png\_path: png/neg\_example.png
\end{tcolorbox}

\begin{tcolorbox}[
  enhanced,
  colback=white,
  colframe=black!20,
  boxrule=0.4pt,
  arc=1mm,
  left=1mm,right=1mm,top=0.6mm,bottom=0.6mm
]
\ttfamily\small
Input (Positive Example): "A bounding box where all text fits fully inside; no crossing or cut off."
\end{tcolorbox}
\begin{tcolorbox}[
  enhanced,
  colback=white,
  colframe=black!12,
  boxrule=0.4pt,
  arc=1mm,
  left=1mm,right=1mm,top=0.6mm,bottom=0.6mm
]
\ttfamily\small
png\_path: png/pos\_example.png
\end{tcolorbox}

\begin{tcolorbox}[
  enhanced,
  colback=white,
  colframe=black!20,
  boxrule=0.4pt,
  arc=1mm,
  left=1mm,right=1mm,top=0.6mm,bottom=0.6mm
]
\ttfamily\small
Input (Target Image): "The final bounding box to classify: overflow vs sparse vs valid."
\end{tcolorbox}
\begin{tcolorbox}[
  enhanced,
  colback=white,
  colframe=black!12,
  boxrule=0.4pt,
  arc=1mm,
  left=1mm,right=1mm,top=0.6mm,bottom=0.6mm
]
\ttfamily\small
png\_path: png/target.png
\end{tcolorbox}

\textbf{Output:}
\begin{itemize}
  \item \texttt{"1"} for overflow
  \item \texttt{"2"} for sparse
  \item \texttt{"3"} for valid
\end{itemize}

\textbf{User Prompt}

Instructions:
\begin{enumerate}
  \item You are provided three images (negative example, positive example, and target).
  \item Refer to the first two images (negative and positive examples) to understand:
  \begin{itemize}
    \item What text overflow looks like
    \item What too much blank space in a bounding box means
    \item How a generally well-fitted bounding box appears
  \end{itemize}
  \item Analyze the third (Target) image's bounding box to check:
  \begin{itemize}
    \item If there is overflow text, return \texttt{"1"}
    \item If there is too much blank space, return \texttt{"2"}
    \item Otherwise (if everything looks good), return \texttt{"3"}
  \end{itemize}
\end{enumerate}

Please analyze the target image and respond with only \texttt{"1"}, \texttt{"2"}, or \texttt{"3"}.

\end{tcolorbox}

\subsection{Prompt of HTML poster code generation}
\begin{tcolorbox}[
  enhanced,
  breakable,
  colback=gray!4,
  colframe=black!45,
  title=\textbf{Prompt: 5-HTML},
  fonttitle=\bfseries,
  boxrule=0.6pt,
  arc=2mm,
  left=1.2mm,right=1.2mm,top=1.0mm,bottom=1.0mm,
  before skip=6pt, after skip=6pt,
]

You are a document-to-poster generation agent. Your task is to read the supplied Markdown text
(\texttt{document\_markdown}) and design a professional, visually appealing academic conference
poster by generating an HTML file. Follow the guidelines below precisely.

\textbf{Instructions}
\begin{enumerate}
  \item Carefully read the Markdown in \texttt{document\_markdown}.
  \item Design a full-page academic conference poster in HTML + CSS:
  \begin{itemize}
    \item Include a prominent header with title, authors, and affiliations. [1ex]
    \item Break content into logical sections (Introduction, Methods, Results, Conclusions, etc.).
    \item Provide clear, informative text summaries.
    \item Embed relevant figures and tables, neatly arranged and aligned.
    \item Accurately represent key findings, methods, and conclusions.
    \item Ensure the layout is engaging, easy to follow, and visually attractive.
    \item Include all essential poster elements commonly found at scientific conferences.
  \end{itemize}
  \item Write complete HTML code (with inline or embedded CSS) that, when rendered, produces the poster layout.
  \item The poster width should be \texttt{poster\_width} px and height should be \texttt{poster\_height} px.
  \item \textbf{Output only} a JSON object with a single key \texttt{"HTML"}, whose value is the entire HTML code for the poster.
\end{enumerate}

\end{tcolorbox}

\section{Guidelines}\label{appendix:guidelines}

\subsection{Poster Construction Guideline for Panel-Based Overflow Annotation} 
\label{app-gud-dataconstr}

\begin{tcolorbox}[
  enhanced,
  breakable,
  colback=gray!4,
  colframe=black!45,
  title=\textbf{Poster Construction Guideline},
  fonttitle=\bfseries,
  boxrule=0.6pt,
  arc=2mm,
  left=1.2mm,right=1.2mm,top=1.0mm,bottom=1.0mm,
  before skip=6pt, after skip=6pt,
  label={box:element}
]

To ensure consistency, realism, and class separability in the manual construction of panel-level poster samples, all contributors must strictly follow the design principles outlined below. These instructions are crafted to guide the generation of high-quality, labelable panels suitable for ternary classification into \texttt{overflow}, \texttt{sparse}, and \texttt{valid}.

\textbf{1. Single-Panel Content Placement:}
Each synthetic poster must contain exactly one panel with content; all other panels should remain fully blank. This isolates the visual characteristics of the target panel and avoids interference from surrounding content.

\textbf{2. Realistic Academic Layout Simulation:}
The filled panel should simulate plausible academic content, including titles, paragraphs, lists, captions, or placeholders for figures and tables. Contributors may adapt excerpts from real research papers or conference posters. Visual realism and domain relevance are essential.

\textbf{3. Manual Composition without Automation:}
All layouts should be manually constructed using basic layout tools. Automated layout engines or template-based generators are prohibited. Contributors must make deliberate layout decisions regarding spacing, font sizing, and alignment.

\textbf{4. Style Diversity and Layout Variation:}
Contributors are encouraged to create diverse samples across multiple layout styles. Variations may include:

\begin{itemize}
    \item Dense vs. sparse textual arrangement
    \item Vertical vs. horizontal alignment preferences
    \item Use or absence of visual anchors (e.g., images, figure boxes)
    \item Differences in title/paragraph proportions

\end{itemize}

\textbf{5. Avoid Intentional Bias toward Label Categories:}
Contributors must avoid consciously designing panels to look obviously “overflow” or “sparse.” Instead, they should focus on authentic academic presentation. The resulting class (e.g., \texttt{overflow}) should arise naturally due to content length and layout tension.

\textbf{6. Respect for Visual Aesthetics:}
Panels should reflect basic academic aesthetics, including reasonable margins, paragraph spacing, and visual balance. Even sparse panels should not appear broken or unprofessional.

\textbf{7. Independence of Construction and Review:}
The annotators responsible for final labeling should not review their own constructed posters. This separation minimizes annotation bias and improves label reliability.

\textbf{8. No Numeric Constraints or Thresholds:}
This guideline does not prescribe any numeric criteria (e.g., font size, character count, margin width). Contributors should rely on visual judgment and academic formatting experience.

Following these principles ensures that the dataset contains structurally diverse, high-quality panels with meaningful visual differences, facilitating rigorous evaluation of overflow detection models.

\end{tcolorbox}

\subsection{Panel Review Guideline for Overflow Detection Annotation}
\label{app-gud-datareview}
\begin{tcolorbox}[
  enhanced,
  breakable,
  colback=gray!4,
  colframe=black!45,
  title=\textbf{Poster Construction Guideline},
  fonttitle=\bfseries,
  boxrule=0.6pt,
  arc=2mm,
  left=1.2mm,right=1.2mm,top=1.0mm,bottom=1.0mm,
  before skip=6pt, after skip=6pt,
  label={box:element}
]

To ensure consistent and unbiased labeling across annotators, each reviewer must follow the structured review protocol outlined below. This guideline is used in conjunction with the poster construction guideline (see Guideline~\ref{app-gud-dataconstr}) to assign one of three layout condition labels to each panel: \texttt{overflow}, \texttt{sparse}, or \texttt{valid}.

\textbf{1. Review Unit:}  
Each review instance consists of a single panel containing content, embedded within a poster template where all other regions are left blank. The task is to judge whether the content visually fits the designated panel appropriately.

\textbf{2. Label Definitions (Visual Semantics):}  
Each panel should be assigned one of the following three labels:
\begin{itemize}
    \item \texttt{overflow}: Content clearly exceeds the visible boundaries of the panel. This may manifest as clipping, truncated text, or compressed elements that breach margins or collide with borders.
    \item \texttt{sparse}: Content occupies only a small fraction of the panel, leading to excessive empty space. The layout appears under-utilized or visually imbalanced.
    \item \texttt{valid}: Content fits comfortably within the panel area with appropriate margins. Neither overcrowded nor excessively empty, the layout appears balanced and professionally acceptable.
\end{itemize}

\textbf{3. Visual Judgment Criteria:}  
Reviewers should rely on human visual perception rather than precise measurements. The following cues are helpful for assessment:
\begin{itemize}
    \item Text proximity to edges or border collisions
    \item Presence of cut-off paragraphs or hidden content
    \item Large central voids or excessive whitespace
    \item Readability and visual comfort
    \item Balance between visual elements (e.g., title, body, figure)
\end{itemize}

\textbf{4. Context-Agnostic Assessment:}  
Labels should be assigned based solely on the visual condition of the current panel, independent of the surrounding poster layout (which is intentionally blank). No domain-specific knowledge or content semantics is required.

\textbf{5. No Use of Quantitative Thresholds:}  
Reviewers must not attempt to count characters, calculate margins, or enforce any numeric thresholds. All judgments are to be made holistically and visually.

\textbf{6. Avoid Overcorrection Bias:}  
Reviewers should not normalize labels across samples. Each panel is to be assessed independently. A valid panel does not require perfect centering or symmetry—just reasonable design aesthetics.

\textbf{7. Consensus via Majority Voting:}  
Each panel is independently reviewed by all five annotators. The final label is determined through majority voting without discussion. In cases of tie or disagreement, the panel is flagged for secondary adjudication (if applicable).

This guideline aims to ensure fair, consistent, and perceptually valid labeling of poster panels for overflow detection evaluation.

\end{tcolorbox}

\subsection{Human Annotation Guideline for Automated Slide Generation}
\label{app-gud-slide-annotation}

\begin{tcolorbox}[
  enhanced,
  breakable,
  colback=gray!4,
  colframe=black!45,
  title=\textbf{Slide Evaluation Guideline},
  fonttitle=\bfseries,
  boxrule=0.6pt,
  arc=2mm,
  left=1.2mm,right=1.2mm,top=1.0mm,bottom=1.0mm,
  before skip=6pt, after skip=6pt,
  label={box:slide-eval}
]

To systematically evaluate the quality of automatically generated presentation slides, we established a comprehensive human annotation framework. To ensure fair, consistent, and unbiased labeling across annotators, each reviewer must follow the structured review protocol outlined below. This guideline defines six weighted evaluation dimensions and a standardized scoring process.

\textbf{1. Review Unit:} Each review instance consists of a complete slide deck generated from a single source paper. The task is to assess the overall quality of the generated slides across multiple dimensions.

\textbf{2. Evaluation Dimensions (Weighted Criteria):} Each slide deck is assessed based on the following weighted criteria:

\centering
\renewcommand{\arraystretch}{1.3}
\resizebox{\linewidth}{!}{\begin{tabular}{clcp{6.8cm}}
\toprule
\textbf{No.} & \textbf{Dimension} & \textbf{Weight} & \textbf{Description} \\
\midrule
1 & Content Completeness & 20\% & Assesses whether the generated slides adequately cover the essential components of the source paper, including abstract, background, methodology, experiments, and conclusions. \\
2 & Logical Structure & 20\% & Examines the rationality of section organization and the coherence of information flow throughout the presentation. \\
3 & Technical Accuracy & 20\% & Evaluates the correctness of domain-specific terminology and the precision of core methodology descriptions. \\
4 & Information Density & 15\% & Measures whether the level of detail is appropriate, avoiding both redundancy and critical omissions. \\
5 & Visual Presentation & 15\% & Considers the integration of figures and text, layout aesthetics, and overall readability. \\
6 & Error Detection & 10\% & Identifies the presence of duplicate sections, content contradictions, or other apparent errors. \\
\bottomrule
\end{tabular}}

\raggedright
\textbf{3. Scoring Criteria:} Reviewers should rely on holistic judgment rather than rigid checklists. The following cues are helpful for assessment:
\begin{itemize}[leftmargin=1.5em, itemsep=0pt, topsep=2pt]
    \item Coverage of key paper sections and contributions
    \item Smooth transitions between slides
    \item Appropriate use of technical terminology
    \item Balance between text and visual elements
    \item Absence of redundant or contradictory content
\end{itemize}

\textbf{4. Independent Assessment:} Scores should be assigned based solely on the quality of the current slide deck, independent of other samples in the evaluation set. No cross-sample normalization is permitted.

\textbf{5. No Use of Quantitative Thresholds:} Reviewers must not attempt to count slides, calculate word counts, or enforce any numeric thresholds. All judgments are to be made holistically based on perceived quality.

\textbf{6. Avoid Overcorrection Bias:} Reviewers should not normalize scores across samples. Each slide deck is to be assessed independently. A high-quality slide deck does not require perfection—just reasonable adherence to academic presentation standards.

\textbf{7. Consensus via Majority Voting:} Each slide deck is independently reviewed by all annotators. The final score for each dimension is determined through averaging or majority voting (for categorical judgments) without discussion. In cases of significant disagreement, the sample is flagged for secondary adjudication (if applicable).

This guideline aims to ensure fair, consistent, and perceptually valid evaluation of automatically generated presentation slides.

\end{tcolorbox}

\end{document}